\pdfoutput=1
\documentclass[11pt]{article}

\usepackage{acl}

\usepackage{times}
\usepackage{latexsym}

\usepackage[T1]{fontenc}

\usepackage[utf8]{inputenc}

\usepackage{microtype}

\usepackage{inconsolata}

\usepackage{graphicx}

\usepackage{tcolorbox}
\usepackage{microtype}
\usepackage{url}
\usepackage{booktabs}
\usepackage{subfigure}

\usepackage{enumitem}
\newenvironment{itemize*}%
 {\leftmargini=20pt\begin{itemize}%
  \setlength{\itemsep}{3pt}%
  \setlength{\parskip}{0pt}%
  }%
 {\end{itemize}} 
\newenvironment{enumerate*}%
 {\begin{enumerate}%
  \setlength{\itemsep}{0pt}%
  \setlength{\parskip}{0pt}}%
 {\end{enumerate}}

\usepackage{amsmath}
\usepackage{cleveref}
\usepackage{listings}
\usepackage{titlesec}
\usepackage{multirow}
\usepackage{makecell}
\usepackage{subcaption} 
\usepackage{xcolor}

\usepackage{array}       
\usepackage{caption}     
\usepackage{amssymb}
\usepackage{dsfont}

\definecolor{midnightgreen}{rgb}{0.0, 0.29, 0.33}
\definecolor{deepgreen}{HTML}{0aa344}
\definecolor{deeppurple}{HTML}{7030a0}
\definecolor{deepblue}{HTML}{171d91}
\definecolor{brown}{HTML}{843c0c}
\definecolor{shadered}{HTML}{ffe5e5}
\definecolor{shadegreen}{HTML}{e5f7ed}
\definecolor{msftBlack}{RGB}{0,0,0}
\definecolor{lightred}{RGB}{255,163,163}
\definecolor{deepred}{RGB}{146,0,0}

\newcommand{\green}{\textcolor{deepgreen}}
\newcommand{\blue}{\textcolor{deepblue}}
\newcommand{\purple}{\textcolor{deeppurple}}

\usepackage[tikz]{bclogo}
\newcommand{\finding}[1]{
  \begin{bclogo}[couleur=msftBlack!05, epBord=1, arrondi=0.2, logo=\bcetoile, marge=5, ombre=true, blur, couleurBord=msftBlack!10, tailleOndu=1, sousTitre={\em #1}]{}
  \end{bclogo}
}

\NewDocumentCommand{\heng}
{ mO{} }{\textcolor{red}{\textsuperscript{\textit{Heng}}\textsf{\textbf{\small[#1]}}}}
\NewDocumentCommand{\cheng}
{ mO{} }{\textcolor{orange}{\textsuperscript{\textit{Cheng}}\textsf{\textbf{\small[#1]}}}}
\NewDocumentCommand{\xiusi}
{ mO{} }{\textcolor{purple}{\textsuperscript{\textit{Xiusi}}\textsf{\textbf{\small[#1]}}}}
\NewDocumentCommand{\hongru}
{ mO{} }{\textcolor{green}{\textsuperscript{\textit{Hongru}}\textsf{\textbf{\small[#1]}}}}
\NewDocumentCommand{\emre}
{ mO{} }{\textcolor{blue}{\textsuperscript{\textit{Emre}}\textsf{\textbf{\small[#1]}}}}

\usepackage[normalem]{ulem}        

\title{ToolRL: Reward is All Tool Learning Needs}

\author{
Cheng Qian, Emre Can Acikgoz, Qi He, Hongru Wang, Xiusi Chen,\\
\textbf{Dilek Hakkani-Tür, Gokhan Tur, Heng Ji}\\
University of Illinois Urbana-Champaign\\
\texttt{\{chengq9, hengji\}@illinois.edu}\\
}

\begin{document}
\maketitle
\begin{abstract}
Current Large Language Models (LLMs) often undergo supervised fine-tuning (SFT) to acquire tool use capabilities. However, SFT struggles to generalize to unfamiliar or complex tool use scenarios. Recent advancements in reinforcement learning (RL), particularly with R1-like models, have demonstrated promising reasoning and generalization abilities. Yet, reward design for tool use presents unique challenges: multiple tools may be invoked with diverse parameters, and coarse-grained reward signals, such as answer matching, fail to offer the finegrained feedback required for effective learning.
In this work, we present the first comprehensive study on reward design for tool selection and application tasks within the RL paradigm. We systematically explore a wide range of reward strategies, analyzing their types, scales, granularity, and temporal dynamics. Building on these insights, we propose a principled reward design tailored for tool use tasks and apply it to train LLMs using Group Relative Policy Optimization (GRPO).
Empirical evaluations across diverse benchmarks demonstrate that our approach yields robust, scalable, and stable training, achieving a 17\% improvement over base models and a 15\% gain over SFT models. These results highlight the critical role of thoughtful reward design in enhancing the tool use capabilities and generalization performance of LLMs. All the code are released to facilitate future research.\footnote{\ Data and codes released at \url{https://github.com/qiancheng0/ToolRL}}
\end{abstract}

\section{Introduction}

\begin{figure}
    \centering
    \includegraphics[width=0.9\linewidth]{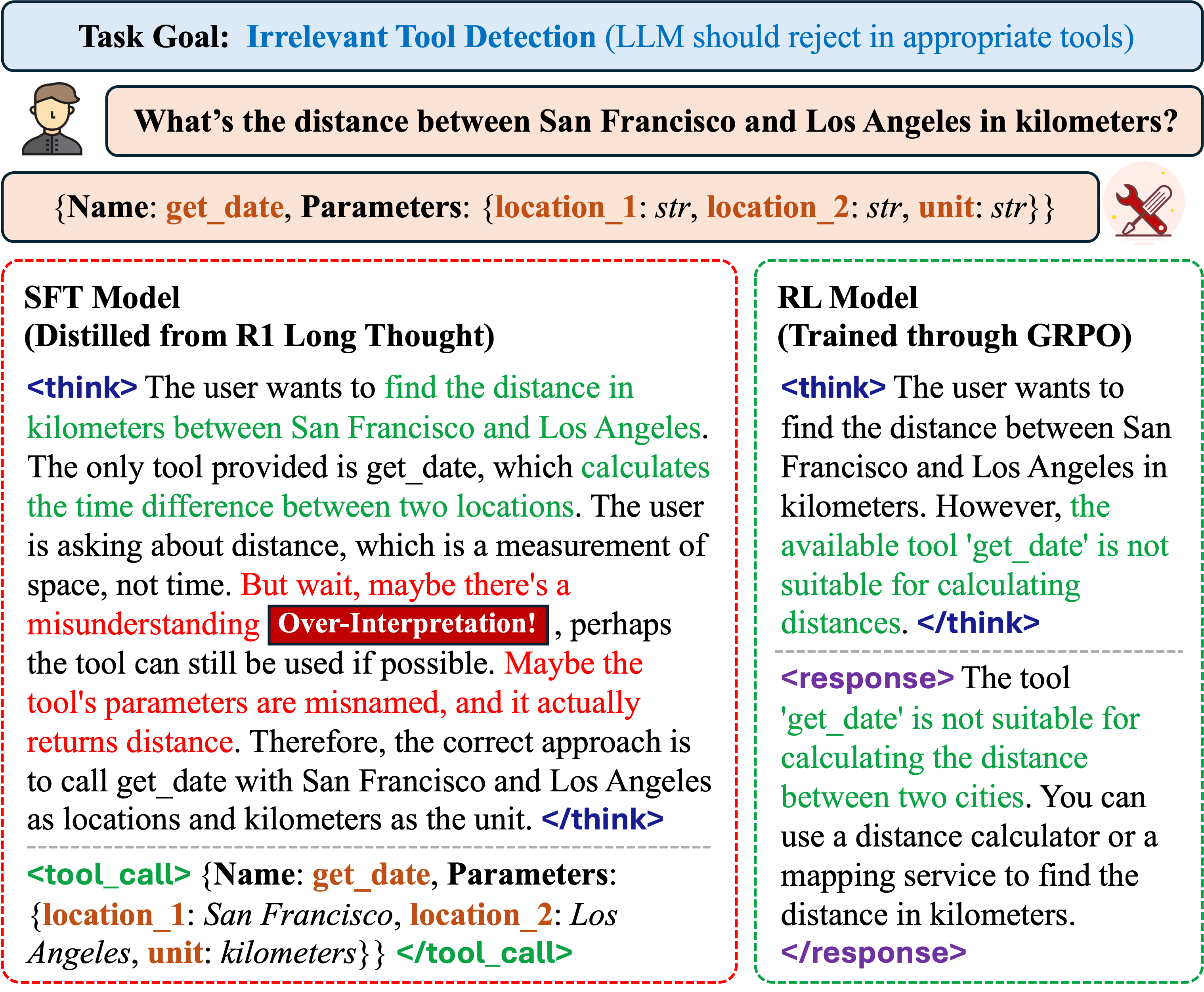}
    \caption{SFT on distilled deep-thinking trajectories suffers from overthinking and limited generalization.} 
    \label{fig:introduction}
\end{figure}


\begin{figure*}
    \centering
    \includegraphics[width=\linewidth]{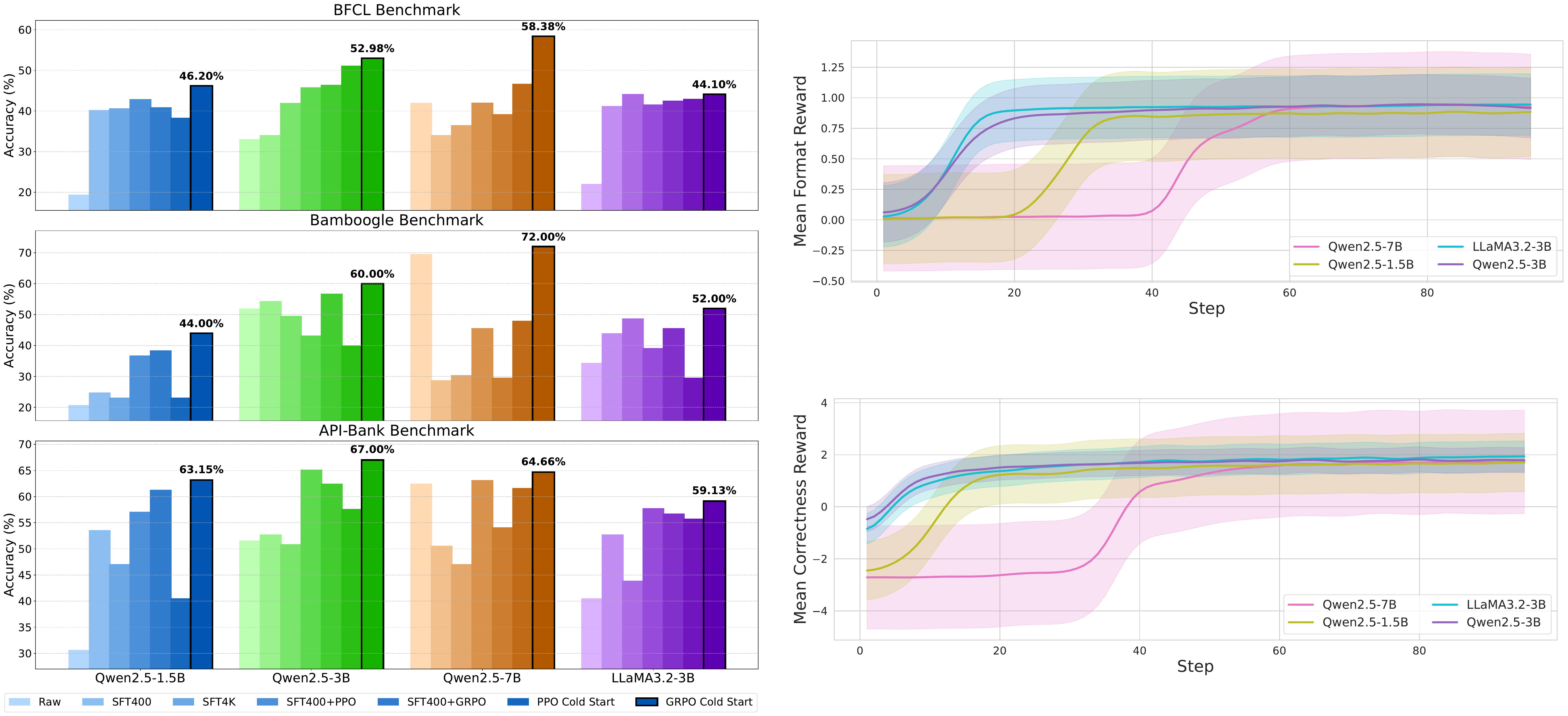}
    \caption{Main results (left) and reward trends over training steps for GRPO Cold Start across four models (right). GRPO Cold Start, equipped with our proposed reward design, consistently achieves the highest performance, with reward curves showing a rapid increase during training.}
    \label{fig:teaser}
\end{figure*}



Recent advances in Large Language Models (LLMs) have showcased remarkable capabilities in complex reasoning tasks~\citep{kumar2025llmreasoningsurvey}. Among the techniques that have significantly contributed to this progress, Reinforcement Learning (RL) has emerged as a powerful paradigm, enabling LLMs to develop emergent capabilities such as self-reflection, self-correction, and long-horizon planning~\citep{guo2025deepseek, team2025kimi}. These capabilities have been instrumental in the success of models like o1 and R1, particularly in mathematical and logical reasoning domains~\citep{qin2024o1, huang2024o1, li2025torl, kang2025t1}.

Beyond traditional reasoning tasks, an increasingly important area is \textbf{Tool-Integrated Reasoning} (TIR). TIR involves LLMs interacting with external tools, such as search engines~\citep{jin2025search, zheng2025deepresearcher}, calculators~\citep{chen2023good, qin2023tool}, or code interpreters~\citep{gou2023tora, liao2024mario}, in a multi-step, feedback-driven loop to arrive at solutions. TIR is particularly important because it addresses core limitations of LLMs, such as outdated knowledge, calculation inaccuracy, and shallow reasoning. By integrating external tools that offer real-time access and specialized capabilities, TIR enables models to tackle complex tasks in a more grounded and goal-directed way. 

Unlike textual reasoning, which primarily involves deduction and inference from static text, TIR additionally demands the model’s ability to select appropriate tools, interpret intermediate outputs, and adaptively refine its trajectory on the fly. These dynamic and interactive reasoning skills position TIR at the core of the emerging paradigm of LLMs-as-agents. As such, TIR enables a wide range of applications, including scientific discovery~\citep{roohani2024biodiscoveryagent, inoue2024drugagent}, research automation~\citep{baek2024researchagent, wang2024autosurvey}, embodied task completion~\citep{zhang2023building, huang2023embodied}, and everyday decision-making~\citep{ye2023rational, zhai2024enhancing}.

Training LLMs for TIR tasks has predominantly relied on Supervised Fine-Tuning (SFT), wherein existing approaches typically generate these integrated reasoning steps offline, followed by subsequent SFT on these trajectories~\citep{chen2023fireact, zeng2024agenttuning, chen2024agentflan, acikgoz2025coalm}. While SFT is effective to some extent, it struggles with generalization, exploration, and adaptability~\citep{chu2025sft, guo2025deepseek}. As illustrated in \Cref{fig:introduction}, a model trained with SFT on deep-thinking trajectories over-interprets the tool and fails to reject the inappropriate tool, merely imitating cues like ``but wait'' without engaging in genuine deep thinking. As such, SFT often fails to capture the strategic flexibility needed for optimal tool use, particularly in open-ended or multi-step settings. This motivates a fundamental research question: \textit{Can RL-based training methods better equip LLMs with agentic tool-using capabilities, and if so, what is the optimal RL design for TIR?}


Recent efforts such as Search-R1~\citep{jin2025search} and TORL~\citep{li2025torl} have begun to explore this direction. However, their focus is narrow: either constrained to search tools in question answering settings or code tools in math problem-solving. In contrast, our work aims to study RL-based training for \textit{general-purpose} tool selection and application, across diverse and complex tool sets with different task types.

For an RL algorithm to be effective, a well-designed \textbf{reward} is essential. Unlike math tasks with a single correct answer, Tool-Integrated Reasoning (TIR) tasks introduce multiple layers of complexity: they often involve multi-step interactions where each turn may require invoking multiple tools, each with carefully specified parameters. Designing effective reward signals to guide learning through this complexity remains an open and underexplored challenge. In this paper, we focus on the problem of reward design for TIR and propose a principled, generalizable framework that can be applied across various RL algorithms. While our reward design is algorithm-agnostic by nature, we empirically demonstrate its effectiveness using both Group Relative Policy Optimization (GRPO)~\citep{shao2024deepseekmath} and Proximal Policy Optimization (PPO)~\citep{schulman2017proximal}, showcasing its versatility and impact on improving tool use performance.

We begin by formalizing the TIR task, and outlining general principles for effective reward design. Building on this foundation, we show how RL algorithm can be leveraged to train LLMs for robust and context-aware tool selection and application. Empirical results demonstrate that our approach outperforms base models by 17\% and SFT models by 15\% across multiple tool use and QA benchmarks. Moreover, the trained model exhibits strong generalization to unseen scenarios and task objectives, along with emergent behaviors such as proactiveness and metacognitive reasoning.

To identify optimal reward strategies, we next systematically explore a broad spectrum of reward configurations across four key dimensions: (1) reward type (what aspect to reward), (2) reward scale (how much to reward), (3) reward granularity (how detailed the reward signal is), and (4) reward dynamics (how rewards evolve over time). Through extensive experiments, we identify reward designs that best align with agentic tool use and uncover insights into what makes a reward ``useful'' for tool invoking LLMs. We summarize the core insights we derive as follows:
\begin{itemize}[topsep=2pt, leftmargin=10pt, itemsep=-2pt]
    \item Longer reasoning trace is not inherently better and length rewards can degrade performance.
    \item Dynamic reward scale helps models transition smoothly from simple to complex behaviors.
    \item Finegrained reward decomposition leads to more stable and effective learning.
\end{itemize}
We also summarize the overall contributions of our paper as follows:
\begin{itemize}[topsep=2pt, leftmargin=10pt, itemsep=-2pt]
    \item We present the first systematic study on RL-based training for general-purpose tool selection and application in LLMs.
    \item We propose a principled reward design framework tailored for TIR and validate its effectiveness through RL algorithms including GRPO.
    \item We conduct extensive experiments analyzing the effects of various reward strategies and distill actionable insights for future research on LLM-agent training.
\end{itemize}
This work pioneers the application of RL to general TIR and provides the first empirical roadmap for reward design in TIR, paving the way toward more capable and autonomous LLM agents.

\section{Related Work}

\paragraph{Tool-Integrated Reasoning of LLMs.}
Tool-integrated reasoning (TIR) has emerged as a promising approach to enhance the capabilities of LLMs. Early studies introduced the concept of equipping LLMs with external tools to overcome their inherent limitations \cite{schick2023toolformer, qin2024tool, yao2023react}, such as program executors \cite{chen2022program} and search engines \cite{vu2023freshllms}. To systematically assess these enhanced capabilities, several benchmarks have been proposed to evaluate tool use performance across various dimensions, including API selection, argument generation, and generalization \cite{qin2023toolllm, patil2023gorilla, qian2024escapebench}. Building on this foundation, subsequent research has focused on constructing high-quality tool use datasets \cite{liu2024toolace, qian2025smart}, enabling models to autonomously create and invoke tools \cite{qian2023creator, qian2024toolink}, and applying these techniques to problems spanning different modalities \cite{shen2025vlm} and specialized domains \cite{ling2023domain}. More recently, reinforcement learning (RL) has been explored as an effective framework to further improve TIR, demonstrating success in tasks such as information retrieval \cite{jin2025search} and math computation \cite{li2025torl}. These advances collectively highlight the growing potential of tool-augmented LLMs for general-purpose reasoning in open-domain settings.

\paragraph{Exploration of RL in LLMs.}
Previous work has primarily relied on supervised fine-tuning (SFT) with carefully curated datasets to enhance LLM performance in tool use~\cite{schick2023toolformer, qin2023toolllm}. Recently, reinforcement learning (RL) has gained traction as a more scalable and generalizable training paradigm. The development of RL methods for LLMs has evolved from reinforcement learning from human feedback (RLHF)~\cite{kaufmann2023survey} and proximal policy optimization (PPO)~\cite{schulman2017proximal} to more advanced techniques such as direct preference optimization (DPO)~\cite{rafailov2023direct}, SimPO~\cite{meng2024simpo}, and group relative policy optimization (GRPO)~\cite{shao2024deepseekmath}. Extensions like dynamic sampling policy optimization (DAPO)~\cite{yu2025dapo} and the more recent value-based augmented proximal policy optimization (VAPO)~\cite{yuan2025vapo} further improve training stability and efficiency.

Among these, GRPO~\cite{shao2024deepseekmath} is specifically designed for LLMs, replacing the traditional critic with a group-based evaluation strategy. It has shown strong performance in enhancing reasoning abilities across a range of tasks, including mathematical problem solving~\cite{shao2024deepseekmath, xie2025logic}, search engine interaction~\cite{jin2025search, song2025r1}, and code generation~\cite{li2025torl}. Beyond task variety, recent studies have analyzed the influence of dataset scale~\cite{li2025limr} and GRPO's effectiveness in smaller model settings~\cite{dang2025reinforcement}. GRPO’s flexible reward function enables adaptation to diverse objectives, such as assigning weights to sub-tasks~\cite{yu2024steptool} or constraining tool use frequency~\cite{li2025torl}. In this work, we extend GRPO to enhance general tool use capabilities, improving LLMs' ability to select and interact with external tools across a wide range of scenarios. 

\section{Method}

Supervised fine-tuning (SFT), as illustrated in \Cref{fig:introduction}, often suffers from overfitting to certain patterns and constrains the model’s ability to learn optimal strategies for tool use. To address this, we introduce a reinforcement learning (RL) approach for enhancing tool-integrated reasoning (TIR) in LLMs. In this section, we begin by defining the TIR task (\Cref{sec:task_definition}), followed by our customized rollout strategy (\Cref{sec:tir_rollout}) and reward design (\Cref{sec:reward_design}). These components are then integrated into the Group Relative Policy Optimization (GRPO) framework~\cite{shao2024deepseekmath} to guide model training on general TIR tasks (\Cref{sec:rl_training}).

\subsection{Task Definition}
\label{sec:task_definition}
\textit{Tool-Integrated Reasoning} (TIR) is the process of incorporating external tools into the reasoning trajectory of an LLM to solve a user task. A typical TIR trajectory involves multiple tool invocations over several reasoning steps, with the final outcome determined by the cumulative success of these intermediate decisions.

Formally, given a tool set \( \mathcal{T} = \{ t_1, t_2, \ldots, t_n \} \) containing \( n \) available tools, and a user query \( Q \), the reasoning trajectory up to step \( k \) is denoted as:
\[
s_k = \left( r_1, \mathcal{T}_1, o_1 \right), \left( r_2, \mathcal{T}_2, o_2 \right), \ldots, \left( r_k, \mathcal{T}_k, o_k \right),
\]
where \( r_i \) denotes the model's natural language reasoning at step \( i \), \( \mathcal{T}_i \subseteq \mathcal{T} \) denotes the set of tool calls invoked at step \( i \), and \( o_i \) denotes the observation received after executing tools in \( \mathcal{T}_i \), possibly including both environment and user feedback.

At each step \( k+1 \), the model must generate the next reasoning step \( r_{k+1} \), select a set of tools \( \mathcal{T}_{k+1} \subseteq \mathcal{T} \), and formulate a grounded tool call (i.e., a parameterized invocation of each tool) to make progress toward solving \( Q \).

The model's policy is defined as \( \pi: s_k \to \left( r_{k+1}, \mathcal{T}_{k+1} \right) \), where the model's objective at each step is to select a tool set \( \mathcal{T}_{k+1} \) that maximizes the immediate reward:
\[
\mathcal{T}_{k+1}^* = \arg\max_{\mathcal{T}_{k+1} \subseteq \mathcal{T}} \; R(s_k, \mathcal{T}_{k+1}, o_{k+1}),
\]
where \( R(\cdot) \) represents the reward function that evaluates progress made by invoking the tools in \( \mathcal{T}_{k+1} \).

While the immediate reward at each step is maximized, the model's policy is implicitly optimized to maximize the cumulative reward over the entire trajectory, formulated as:
\[
\max_\pi \; \mathbb{E}_{\pi} \left[ \sum_{k=1}^{K} R(s_k, \mathcal{T}_{k+1}, o_{k+1}) \right],
\]
This formulation is valid because our training data includes ground truth tool calls at each step, allowing step-wise reward signals to guide multi-step success. Unlike QA tasks that focus solely on the final answer, tool selection and application tasks provide dense intermediate feedback. Moreover, we later demonstrate that our method enables the model to generalize to settings where tool calls are free-form and only the final outcome matters. Therefore, out task setting encourages the model to optimize tool use at each step while aligning with the overall task goal.




\begin{figure*}
    \centering
    \includegraphics[width=\linewidth]{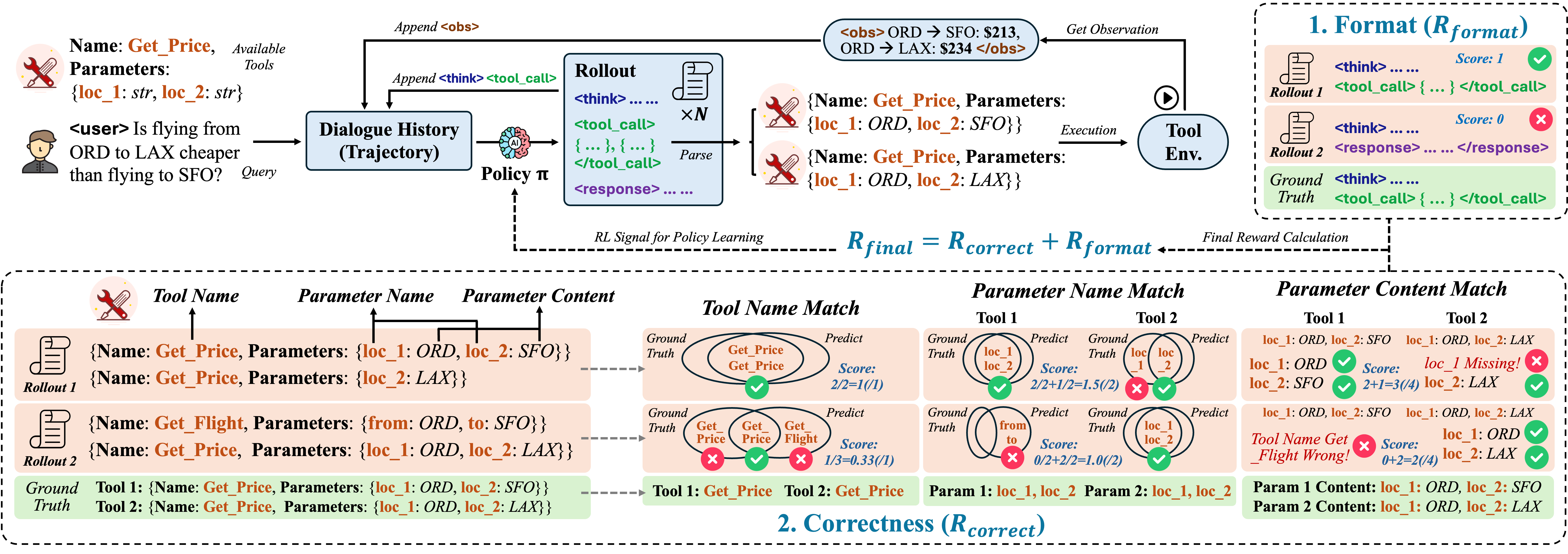}
    \caption{Illustration of TIR rollout and calculation of format and correctness reward.}
    \label{fig:reward}
\end{figure*}


\begin{figure*}[t]
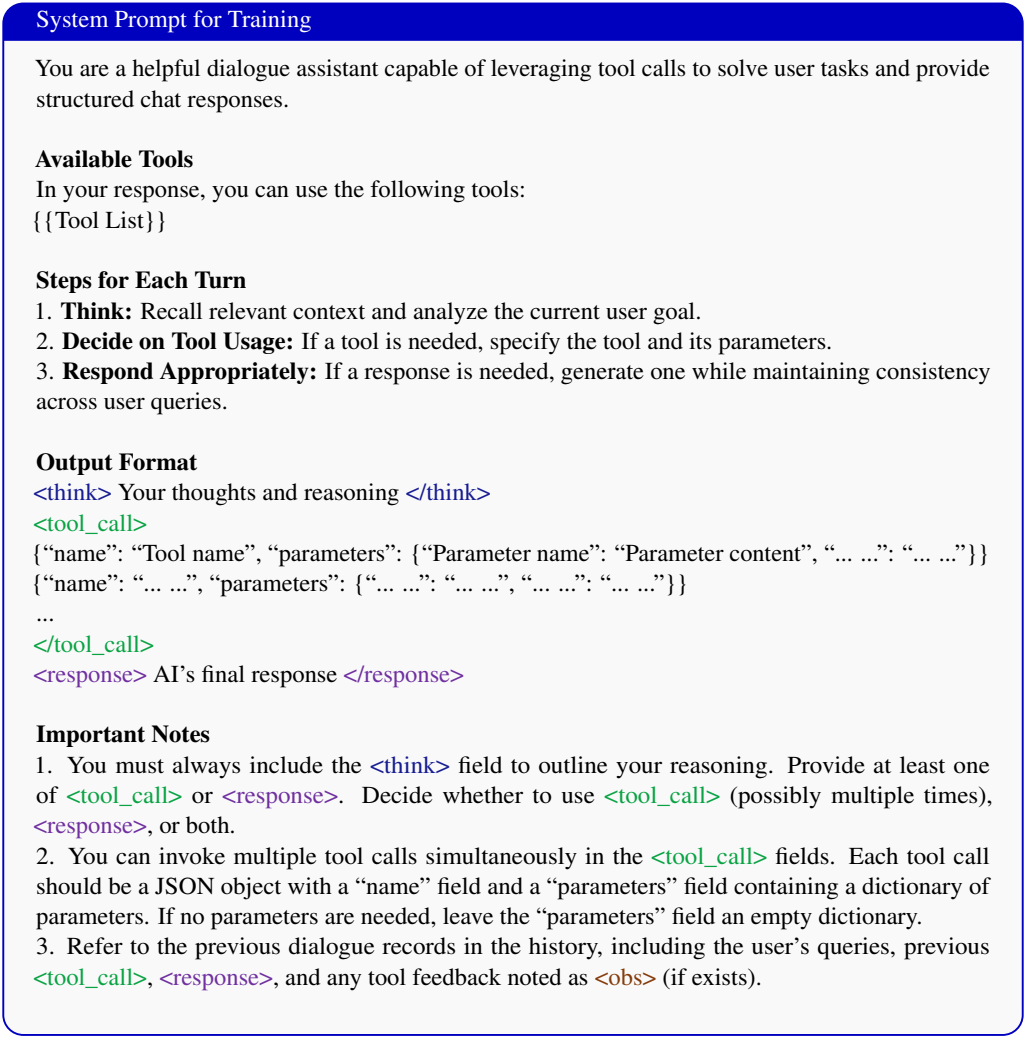

\centering
\resizebox{0.85\textwidth}{!}{
\begin{tcolorbox}[colback=gray!5!white, colframe=blue!75!black, 
title=System Prompt for Training, boxrule=0.3mm, width=\textwidth, arc=3mm, auto outer arc=true]
You are a helpful dialogue assistant capable of leveraging tool calls to solve user tasks and provide structured chat responses.\\
\\
\textbf{Available Tools}\\
In your response, you can use the following tools:\\
\{\{Tool List\}\}\\
\\
\textbf{Steps for Each Turn}\\
1. \textbf{Think:} Recall relevant context and analyze the current user goal.\\
2. \textbf{Decide on Tool Usage:} If a tool is needed, specify the tool and its parameters.\\
3. \textbf{Respond Appropriately:} If a response is needed, generate one while maintaining consistency across user queries.\\
\\
\textbf{Output Format}\\
\textcolor{deepblue}{\textless{}think\textgreater{}} Your thoughts and reasoning \textcolor{deepblue}{\textless{}/think\textgreater{}}\\
\textcolor{deepgreen}{\textless{}tool\_call\textgreater{}}\\
\{``name'': ``Tool name'', ``parameters'': \{``Parameter name'': ``Parameter content'', ``... ...'': ``... ...''\}\}\\
\{``name'': ``... ...'', ``parameters'': \{``... ...'': ``... ...'', ``... ...'': ``... ...''\}\}\\
...\\
\textcolor{deepgreen}{\textless{}/tool\_call\textgreater{}}\\
\textcolor{deeppurple}{\textless{}response\textgreater{}} AI's final response \textcolor{deeppurple}{\textless{}/response\textgreater{}}\\
\\
\textbf{Important Notes}\\
1. You must always include the \textcolor{deepblue}{\textless{}think\textgreater{}} field to outline your reasoning. Provide at least one of \textcolor{deepgreen}{\textless{}tool\_call\textgreater{}} or \textcolor{deeppurple}{\textless{}response\textgreater{}}. Decide whether to use \textcolor{deepgreen}{\textless{}tool\_call\textgreater{}} (possibly multiple times), \textcolor{deeppurple}{\textless{}response\textgreater{}}, or both.\\
2. You can invoke multiple tool calls simultaneously in the \textcolor{deepgreen}{\textless{}tool\_call\textgreater{}} fields. Each tool call should be a JSON object with a ``name'' field and a ``parameters'' field containing a dictionary of parameters. If no parameters are needed, leave the ``parameters'' field an empty dictionary.\\
3. Refer to the previous dialogue records in the history, including the user's queries, previous \textcolor{deepgreen}{\textless{}tool\_call\textgreater{}}, \textcolor{deeppurple}{\textless{}response\textgreater{}}, and any tool feedback noted as \textcolor{brown}{\textless{}obs\textgreater{}} (if exists).\\
\end{tcolorbox}
}
\caption{The system prompt used for TIR's rollout.}
\label{prompt:system}
\end{figure*}


\subsection{TIR Rollout}
\label{sec:tir_rollout}
To enable the model to autonomously generate reasoning traces and tool calls, we utilize a system prompt as shown in \Cref{prompt:system} during rollout. The \textit{Tool List} placeholder denotes the tool set \(\mathcal{T}\), which contains all tools available for invocation. We indicate in the instruction that the LLM should use special tokens \textcolor{deepblue}{\textless{}think\textgreater{}}, \textcolor{deepgreen}{\textless{}tool\_call\textgreater{}}, and \textcolor{deeppurple}{\textless{}response\textgreater{}} to indicates their thoughts, tool calls and responses in output.

As illustrated in \Cref{fig:reward}, when the model output includes \textcolor{deepgreen}{\textless{}tool\_call\textgreater{}}, we automatically parse the tool calls into individual invocations using the model-predicted parameters. The outputs from executions are then inserted into the \textcolor{brown}{\textless{}obs\textgreater{}} field and appended to the dialogue history, whose format is shown in \Cref{prompt:user}, serving as the model's interaction trajectory. Similarly, if the output contains \textcolor{deeppurple}{\textless{}response\textgreater{}}, the corresponding response is parsed and appended to the dialogue history.

It is important to note that \textcolor{deepgreen}{\textless{}tool\_call\textgreater{}} and \textcolor{deeppurple}{\textless{}response\textgreater{}} are not mutually exclusive; they may co-occur within a single output. The user's initial query \( Q \) is placed in the \textit{Initial User Input} placeholder, and any subsequent user inputs are also appended to the dialogue history when present.


\subsection{Reward Design}
\label{sec:reward_design}
Rule-based reward mechanisms have demonstrated strong empirical performance and are commonly employed. In our training, we similarly adopt a reward formulation that combines structural and correctness-based components, in line with prior works~\citep{jin2025search, li2025torl, xie2025logic}. Specifically, the format reward assesses whether the model output adheres to the expected structure including thoughts, tool calls, and responses, while the correctness reward evaluates the accuracy of tool invocations. Formally, the overall reward \( R_{\text{final}}(\cdot) \) is decomposed into two components: \( R_{\text{format}} + R_{\text{correct}} \), each described in detail below:

\paragraph{Format Reward.} The format reward \( \mathcal{R}_{\text{format}} \in \{0, 1\} \) checks whether the model output contains all required special tokens in the correct order as specified by the ground truth:

\begin{equation*}
\mathcal{R}_{\text{format}} = 
\begin{cases}
1, & \vcenter{\hbox{\parbox{5.5cm}{if all required fields appear\\and are in the correct order}}} \vspace{8pt}\\
0, & \text{otherwise}
\end{cases}
\end{equation*}

\paragraph{Correctness Reward.} The correctness reward \( \mathcal{R}_{\text{correct}} \in [-3, 3] \) evaluates predicted tool calls \( P = \{P_1, ..., P_m\} \) against ground-truth calls \( G = \{G_1, ..., G_n\} \). It includes three components:

\begin{itemize}[topsep=2pt, leftmargin=10pt, itemsep=0pt]
\item \textit{Tool Name Matching:}
\begin{small}
\begin{equation*}
r_{\text{name}} = \frac{|N_G \cap N_P|}{|N_G \cup N_P|} \in [0, 1]
\end{equation*}
\end{small}
where \( N_G \) and \( N_P \) are the sets of tool names extracted from the ground-truth and predicted tool calls, respectively. 

\item \textit{Parameter Name Matching:}
\begin{small}
\begin{equation*}
r_{\text{param}} = \sum_{G_j \in G} \frac{|\text{keys}(P_G) \cap \text{keys}(P_P)|}{|\text{keys}(P_G) \cup \text{keys}(P_P)|} \in [0, |G|]
\end{equation*}
\end{small}
where \( \text{keys}(P_G) \) and \( \text{keys}(P_P) \) represent the parameter names of the predicted and ground-truth tool calls, respectively.

\item \textit{Parameter Content Matching:}
\begin{small}
\begin{equation*}
\begin{split}
r_{\text{value}} = \sum_{G_j \in G} \ \sum_{k \in \text{keys}(G_j)} \mathds{1}[P_G[k] = P_P[k]] \\
\in [0, \sum_{G_j \in G} |\text{keys}(G_j)|]
\end{split}
\end{equation*}
\end{small}
where \( P_G[k]] \) and \( P_P[k] \) represent the values of the parameters for the predicted and ground truth tool calls.

\item Total match score for each match is:
\begin{small}
\begin{equation*}
r_{\text{match}} = r_{\text{name}} + r_{\text{param}} + r_{\text{value}} \in [0, S_{\max}]
\end{equation*}
\end{small}
where \( S_{\max} = 1 + |G| + \sum_{G_j \in G} |\text{keys}(G_j)|\) denotes the maximum possible score.
\end{itemize}

\vspace{5pt} \noindent
The total score is computed by finding the optimal matching between \( P \) and \( G \) to maximize the total match score:
\begin{equation*}
\mathcal{R}_{\text{correct}} = 6 \cdot \frac{R_{\max}}{S_{\max}} - 3 \in [-3, 3]
\end{equation*}
where \( R_{\max} \) denotes the total match score from the optimal matching. The final correctness reward \( \mathcal{R}_{\text{correct}} \) is the normalized reward for the matching process. We empirically set the reward scale within the range of \( [-3, 3] \), with more analysis and ablatiions of reward scale presented in \Cref{sec:analysis}.

The final reward value \( \mathcal{R}_{\text{final}} \) is finally derived as the sum of \( \mathcal{R}_{\text{format}} \) and \( \mathcal{R}_{\text{correct}} \):
\[
\mathcal{R}_{\text{final}} = \mathcal{R}_{\text{format}} + \mathcal{R}_{\text{correct}} \in [-3, 4]
\]

Unlike prior works that often rely on binary or overly simplified reward signals, our design captures the nuanced structure of tool calls by evaluating multiple interdependent components including tool names, parameter schemas, and parameter values. This fine-grained formulation better reflects the complexity of real-world tool use, where correctness cannot be reduced to a single binary criterion. We further validate the impact of this design through comprehensive analysis in \Cref{sec:analysis}.

Overall, our reward design ensures a balanced and interpretable evaluation signal by explicitly separating structural compliance from semantic correctness. By aligning rewards with both format adherence and fine-grained tool call accuracy, the model is guided to produce outputs that are not only syntactically valid but also semantically faithful, which is crucial for downstream tool execution and final task success.


\subsection{RL Training with GRPO}
\label{sec:rl_training}
To tune the model with structured rewards, we employ GRPO, a variant of PPO that introduces advantage normalization within grouped samples. This normalization helps stabilize training by reducing variance across samples that share a common input context. Let \(\pi_\theta\) represent the current policy.

\paragraph{Normalized Advantage Across Query Groups.}
For each query \( Q \), its responses derived from the rollout form a group \( G_Q \) consisting of multiple responses and their corresponding reward values: 
\[
G_Q = \{ A, (s_1, r_1), (s_2, r_2), \ldots, (s_n, r_n) \}
\]
where \( A \) denotes the ground-truth annotation for \( Q \), and each reward \( r_i \) is computed as the sum of the format and correctness rewards associated with response \( s_i \), i.e., \( r_i = \mathcal{R}_{\text{format}}(s_i, A) + \mathcal{R}_{\text{correct}}(s_i, A) \). For each group, we calculate the mean and standard deviation of the rewards:

\begin{small}
\begin{equation*}
    \mu_Q = \frac{1}{n} \sum_{i=1}^n r_i, \quad 
    \sigma_Q = \sqrt{\frac{1}{n} \sum_{i=1}^n (r_i - \mu_Q)^2}
\end{equation*}
\end{small}

Then, for each sample \( s_i \) in the group, we define the normalized advantage:

\begin{small}
\begin{equation*}
    A_i(s_i | Q) = \frac{r_i - \mu_Q}{\sigma_Q + \eta}
\end{equation*}
\end{small}
where \( \eta \) is a constant to avoid division by zero.

\paragraph{Policy Optimization Objective.}
The policy \( \pi_\theta \) is optimized using the standard clipped PPO objective, adapted with our group-wise normalized advantages:

\begin{small}
\begin{equation*}
\begin{split}
J_{\text{GRPO}}(\theta) = \mathbb{E}_{Q \sim \mathcal{D}} \mathbb{E}_{s_i \sim \pi_\theta} \Big[ 
\min\Big( \frac{\pi_\theta(s_i|Q)}{\pi_{\text{old}}(s_i|Q)} A_i(s_i|Q), \\
\text{clip}\big( \frac{\pi_\theta(s_i|Q)}{\pi_{\text{old}}(s_i|Q)}, 1-\epsilon, 1+\epsilon \big) A_i(s_i|Q) 
\Big) \Big]
\end{split}
\end{equation*}
\end{small}

Unlike the original GRPO formulations, we omit the KL penalty term against a reference model. This design choice encourages the model to more freely adapt its behavior to our custom response format and structured reward signals. In practice, we observe that this leads to faster convergence and comparable performance, while also simplifying the training pipeline.

Overall, this objective guides the policy to generate structurally consistent and semantically accurate tool calls, while group-wise normalization mitigates reward variance across queries, leading to more stable and sample-efficient alignment with task-specific response requirements.

\section{Experiments}

\subsection{Training Dataset}
To support robust tool learning through RL, we construct a mixed dataset spanning diverse tool use scenarios:

\begin{itemize}[topsep=2pt, leftmargin=10pt, itemsep=-2pt]
\item \textbf{ToolACE}~\citep{liu2024toolace}: A general tool use dataset where the model learns when to invoke tools versus respond directly, improving decision-making in multi-step interactions.

\item \textbf{Hammer (Masked)}~\citep{lin2024hammer}: A subset of Hammer with randomized tool and parameter names, forcing the model to rely on descriptions rather than memorized labels, thus enhancing generalization and reducing overfitting to certain tools.

\item \textbf{xLAM}~\citep{zhang2024xlam}: A compositional dataset requiring one or multiple tool calls per turn, encouraging the model to reason about tool dependencies and plan diverse tool calling action actively.
\end{itemize}

For RL training, we sample 2K examples from ToolACE and 1K each from Hammer and xLAM, creating a balanced dataset spanning diverse levels of complexity and tool use. Multi-step trajectories are decomposed into single-step instances, with prior dialogue history injected into the user prompt (as shown in \Cref{prompt:user}) to preserve context. This setup encourages strategic exploration and teaches the model to select and apply tools appropriately within each step. Please see \Cref{apdx:training} for more details and justifications.

\subsection{Experiment Settings}
\paragraph{Training.}  We conduct all RL experiments using the veRL framework~\citep{sheng2024hybridflow}, adopting the GRPO algorithm detailed in the previous section. For each training step, we sample a batch of 512, and generate 4 responses per query, training for 15 epochs in total (see \Cref{apdx:training} for full configuration details). To encourage broader policy exploration, we remove KL regularization and apply a generation temperature of 1.0. We initialize our models with the Qwen-2.5-Instruct~\citep{2024qwen2.5} and Llama-3.2-Instruct~\citep{dubey2024llama} series, which are further tuned under the GRPO objective with our customized reward design.

\paragraph{Evaluation.} We evaluate our approach on the \textbf{Berkeley Function Call Leaderboard} (BFCL)~\citep{patil2024gorilla}, a comprehensive benchmark that spans a diverse set of challenges, including single-step reasoning, multi-step tool use, real-time execution, irrelevant tool rejection, simultaneous multi-tool selection, and multi-tool application\footnote{\url{https://gorilla.cs.berkeley.edu/blogs/13_bfcl_v3_multi_turn.html}}. In addition, we present results on \textbf{API-Bank}~\citep{li2023api}, a three-level evaluation framework comprising 73 diverse and complex API tools. It assesses an LLM’s ability to select and apply tools through natural multi-turn dialogues, across three levels of difficulty. We also evaluate on a representative QA benchmark \textbf{Bamboogle}~\citep{press2022measuring}, which comprises a variety of question-answering tasks where performance is measured based on the final answer accuracy rather than the correctness of tool use. These broad coverage makes our evaluation setting effective for evaluating real-world LLM tool use proficiency. All results are reported in terms of accuracy.

\paragraph{Baselines.} We compare our approach against several baselines to better isolate the effects of GRPO training: (1) \textbf{Raw Instruct Model}: the original model without any additional fine-tuning or RL, evaluated using the same prompts. (2) \textbf{SFT on RL Data}: the instruct model fine-tuned using the same 4K / selected 400 data points as the RL training set, providing a comparison point to assess whether GRPO training outperforms standard SFT. (3) \textbf{GRPO on SFT Model}: GRPO is applied to a model that has already undergone SFT on the selected 400 data points. This setup allows us to evaluate the impact of initializing GRPO with a format-aware model, in contrast to starting from the raw instruct model in a cold start manner. (4) \textbf{PPO:} We also include the standard PPO setting as a baseline to evaluate whether our reward design is effective beyond GRPO. We report results for both a cold start PPO model and a PPO model initialized with SFT, using the same hyperparameters as in the GRPO setup for a fair comparison. Please refer to \Cref{apdx:training} for more details and justifications.


\subsection{Results}

\begin{table*}[!t]
\centering
\resizebox{1.0\linewidth}{!}{
\begin{tabular}{lccccccc}
\toprule
\textbf{Model} & \textbf{Overall Acc} & Non-Live AST Acc & Non-Live Exec Acc & Live Acc & Multi Turn Acc & Relevance Detection & Irrelevance Detection \\
\midrule
Qwen2.5-1.5B-Instruct (\textbf{Raw}) & 19.41\% & 16.00\% & 13.18\% & 35.58\% & 0.00\% & 44.44\% & 82.49\% \\
Qwen2.5-1.5B-Instruct (\textbf{SFT400}) & 40.21\% & 65.12\% & 61.11\% & 56.69\% & 1.00\% & 94.44\% & 60.14\% \\
Qwen2.5-1.5B-Instruct (\textbf{SFT4k}) & 40.67\% & 59.94\% & 59.84\% & 59.31\% & 1.00\% & 88.89\% & 71.34\% \\
Qwen2.5-1.5B-Instruct (\textbf{SFT400+PPO}) & \uline{42.95\%} & 77.65\% & 69.75\% & 55.73\% & 1.88\% & 100.00\% & 48.40\% \\
Qwen2.5-1.5B-Instruct (\textbf{SFT400+GRPO}) & 40.93\% & 70.54\% & 60.79\% & 56.33\% & 1.00\% & 94.44\% & 58.63\% \\
Qwen2.5-1.5B-Instruct (\textbf{PPO Cold Start}) & 38.32\% & 79.40\% & 70.11\% & 45.24\% & 0.87\% & 100.00\% & 18.09\% \\
Qwen2.5-1.5B-Instruct (\textbf{Ours, GRPO Cold Start}) & \textbf{46.20\%} & 77.96\% & 76.98\% & 60.73\% & 2.25\% & 100.00\% & 56.44\% \\
\midrule
Qwen2.5-3B-Instruct (\textbf{Raw}) & 33.04\% & 42.52\% & 40.80\% & 53.96\% & 1.00\% & 64.71\% & 56.01\% \\
Qwen2.5-3B-Instruct (\textbf{SFT400}) & 34.08\% & 69.29\% & 61.50\% & 41.40\% & 0.00\% & 94.44\% & 8.11\% \\
Qwen2.5-3B-Instruct (\textbf{SFT4k}) & 41.97\% & 62.85\% & 54.73\% & 59.17\% & 0.75\% & 77.78\% & 75.12\% \\
Qwen2.5-3B-Instruct (\textbf{SFT400+PPO}) & 45.80\% & 78.29\% & 71.09\% & 58.76\% & 5.12\% & 94.12\% & 54.70\% \\
Qwen2.5-3B-Instruct (\textbf{SFT400+GRPO}) & 46.42\% & 76.21\% & 68.93\% & 64.15\% & 1.75\% & 88.89\% & 71.76\% \\
Qwen2.5-3B-Instruct (\textbf{PPO Cold Start}) & \uline{51.15\%} & 82.42\% & 78.52\% & 67.78\% & 4.88\% & 94.12\% & 73.87\% \\
Qwen2.5-3B-Instruct (\textbf{Ours, GRPO Cold Start}) & \textbf{52.98\%} & 81.58\% & 79.43\% & 73.78\% & 3.75\% & 88.24\% & 84.85\% \\
\midrule
Qwen2.5-7B-Instruct (\textbf{Raw}) & 41.97\% & 66.02\% & 70.11\% & 53.51\% & 4.25\% & 76.47\% & 62.66\% \\
Qwen2.5-7B-Instruct (\textbf{SFT400}) & 34.08\% & 69.29\% & 66.68\% & 41.4\% & 0.00\% & 94.44\% & 8.11\% \\
Qwen2.5-7B-Instruct (\textbf{SFT4k}) & 36.53\% & 45.15\% & 53.5\% & 57.13\% & 0.75\% & 72.22\% & 72.32\% \\
Qwen2.5-7B-Instruct (\textbf{SFT400+PPO}) & 42.02\% & 83.90\% & 72.62\% & 51.84\% & 0.25\% & 100.00\% & 29.66\% \\
Qwen2.5-7B-Instruct (\textbf{SFT400+GRPO}) & 39.25\% & 80.69\% & 74.34\% & 46.51\% & 0.25\% & 100.00\% & 14.19\% \\
Qwen2.5-7B-Instruct (\textbf{PPO Cold Start}) & \uline{46.68\%} & 79.33\% & 78.16\% & 63.17\% & 0.38\% & 88.89\% & 52.92\% \\
Qwen2.5-7B-Instruct (\textbf{Ours, GRPO Cold Start}) & \textbf{58.38\%} & 86.17\% & 78.25\% & 74.9\% & 18.12\% & 83.33\% & 76.68\%  \\
\midrule
Llama-3.2-3B-Instruct (\textbf{Raw}) & 22.09\% & 17.44\% & 14.57\% & 43.85\% & 0.00\% & 77.78\% & 66.07\% \\
Llama-3.2-3B-Instruct (\textbf{SFT400}) & 41.22\% & 64.27\% & 62.18\% & 58.37\% & 0.75\% & 66.67\% & 71.12\% \\
Llama-3.2-3B-Instruct (\textbf{SFT4k}) & \textbf{44.16\%} & 65.42\% & 67.02\% & 63.04\% & 1.38\% & 77.78\% & 78.25\% \\
Llama-3.2-3B-Instruct (\textbf{SFT400+PPO}) & 41.62\% & 68.10\% & 69.88\% & 52.98\% & 3.00\% & 94.12\% & 56.29\% \\
Llama-3.2-3B-Instruct (\textbf{SFT400+GRPO}) & 42.54\% & 65.15\% & 68.98\% & 59.40\% & 0.88\% & 72.22\% & 65.80\% \\
Llama-3.2-3B-Instruct (\textbf{PPO Cold Start}) & 42.98\% & 84.00\% & 72.00\% & 52.80\% & 2.88\% & 100.00\% & 31.94\% \\
Llama-3.2-3B-Instruct (\textbf{Ours, GRPO Cold Start}) & \uline{44.10\%} & 74.38\% & 75.18\% & 56.86\% & 1.37\% & 94.44\% & 62.23\% \\
\bottomrule
\end{tabular}
}
\caption{BFCL V3 Benchmark Results (Main Result)}
\label{tab:bfcl-result-main}
\end{table*}


\begin{table*}[!t]
\centering

\begin{minipage}[t]{0.51\linewidth}
\centering

\resizebox{1\linewidth}{!}{
\begin{tabular}{lcccc}
\toprule
\textbf{Model} & \textbf{Overall Acc} & Level 1 & Level 2 & Level 3 \\
\midrule
Qwen2.5-1.5B-Instruct (\textbf{Raw}) & 30.65\% & 28.32\% & 35.82\% & 35.11\% \\
Qwen2.5-1.5B-Instruct (\textbf{SFT400}) & 53.60\% & 57.14\% & 50.75\% & 44.27\% \\
Qwen2.5-1.5B-Instruct (\textbf{SFT4k}) & 47.07\% & 52.88\% & 52.24\% & 26.72\% \\
Qwen2.5-1.5B-Instruct (\textbf{SFT400+PPO}) & 57.12\% & 60.9\% & 50.75\% & 48.85\% \\
Qwen2.5-1.5B-Instruct (\textbf{SFT400+GRPO}) & \uline{61.31\%} & 64.16\% & 58.21\% & 54.20\% \\
Qwen2.5-1.5B-Instruct (\textbf{PPO Cold Start}) & 40.54\% & 44.61\% & 31.34\% & 32.82\% \\
Qwen2.5-1.5B-Instruct (\textbf{Ours, GRPO Cold Start}) & \textbf{63.15}\% & 70.68\% & 61.19\% & 41.22\% \\
\midrule
Qwen2.5-3B-Instruct (\textbf{Raw}) & 51.59\% & 59.65\% & 32.84\% & 36.64\% \\
Qwen2.5-3B-Instruct (\textbf{SFT400}) & 52.76\% & 59.65\% & 50.75\% & 32.82\% \\
Qwen2.5-3B-Instruct (\textbf{SFT4k}) & 50.92\% & 55.64\% & 43.28\% & 40.46\% \\
Qwen2.5-3B-Instruct (\textbf{SFT400+PPO}) & \uline{65.16\%} & 67.92\% & 55.22\% & 61.83\% \\
Qwen2.5-3B-Instruct (\textbf{SFT400+GRPO}) & 62.48\% & 68.67\% & 58.21\% & 45.80\% \\
Qwen2.5-3B-Instruct (\textbf{PPO Cold Start}) & 57.62\% & 64.66\% & 59.70\% & 35.11\% \\
Qwen2.5-3B-Instruct (\textbf{Ours, GRPO Cold Start}) & \textbf{67.00\%} & 73.43\% & 67.16\% & 47.33\% \\
\midrule
Qwen2.5-7B-Instruct (\textbf{Raw}) & 62.48\% & 70.68\% & 49.25\% & 44.27\% \\
Qwen2.5-7B-Instruct (\textbf{SFT400}) & 50.59\% & 55.89\% & 50.75\% & 34.35\% \\
Qwen2.5-7B-Instruct (\textbf{SFT4k}) & 47.07\% & 51.13\% & 34.33\%&41.22\% \\
Qwen2.5-7B-Instruct (\textbf{SFT400+PPO}) & \uline{63.15\%} & 72.43\% & 58.21\% & 37.40\% \\
Qwen2.5-7B-Instruct (\textbf{SFT400+GRPO}) & 54.10\% & 61.40\% & 52.24\% &32.82\% \\
Qwen2.5-7B-Instruct (\textbf{PPO Cold Start}) & 61.64\% & 68.67\% & 44.78\% & 48.85\% \\
Qwen2.5-7B-Instruct (\textbf{Ours, GRPO Cold Start}) & \textbf{64.66\%} & 73.93\% & 61.19\% & 38.17\% \\
\midrule
Llama-3.2-3B-Instruct (\textbf{Raw}) & 40.54\% & 44.86\% & 29.85\% & 32.82\% \\
Llama-3.2-3B-Instruct (\textbf{SFT400}) & 52.76\% & 60.65\% & 35.82\% & 37.40\% \\
Llama-3.2-3B-Instruct (\textbf{SFT4k}) & 43.89\% & 53.88\% & 29.85\% & 20.61\% \\
Llama-3.2-3B-Instruct (\textbf{SFT400+PPO}) & 57.79\% & 63.16\% & 47.76\% & 46.56\% \\
Llama-3.2-3B-Instruct (\textbf{SFT400+GRPO}) & \uline{56.78\%} & 63.60\% & 41.79\% & 43.51\% \\
Llama-3.2-3B-Instruct (\textbf{PPO Cold Start}) & 55.78\% & 60.65\% & 41.79\% & 48.09\% \\
Llama-3.2-3B-Instruct (\textbf{Ours, GRPO Cold Start}) & \textbf{59.13\%} & 65.66\% & 52.24\% & 42.75\% \\
\bottomrule
\end{tabular}
}
\caption{API-Bank Test Results (Main Result)}
\label{tab:apibank-result-main}

\end{minipage}
\hfill
\begin{minipage}[t]{0.452\textwidth}
\centering

\resizebox{1\linewidth}{!}{
\begin{tabular}{lcccc}
\toprule
\textbf{Model} & \textbf{Accuracy} & Avg Num Tool Call \\
\midrule
Qwen2.5-1.5B-Instruct (\textbf{Raw}) & 20.8\% & 0.61 \\
Qwen2.5-1.5B-Instruct (\textbf{SFT400}) & 24.8\% & 0.78 \\
Qwen2.5-1.5B-Instruct (\textbf{SFT4k}) & 23.2\% & 1.25 \\
Qwen2.5-1.5B-Instruct (\textbf{SFT400+PPO}) & 36.8\% & 1.06 \\
Qwen2.5-1.5B-Instruct (\textbf{SFT400+GRPO}) & \uline{38.4\%} & 0.96 \\
Qwen2.5-1.5B-Instruct (\textbf{PPO Cold Start}) & 23.2\% & 2.38 \\
Qwen2.5-1.5B-Instruct (\textbf{Ours, GRPO Cold Start}) & \textbf{44.0\%} & 1.19 \\
\midrule
Qwen2.5-3B-Instruct (\textbf{Raw}) & 52.0\% & 1.77 \\
Qwen2.5-3B-Instruct (\textbf{SFT400}) & 54.4\% & 0.86 \\
Qwen2.5-3B-Instruct (\textbf{SFT4k}) & 49.6\% & 0.92 \\
Qwen2.5-3B-Instruct (\textbf{SFT400+PPO}) & 43.2\% & 1.04 \\
Qwen2.5-3B-Instruct (\textbf{SFT400+GRPO}) & \uline{56.8\%} & 0.99 \\
Qwen2.5-3B-Instruct (\textbf{PPO Cold Start}) & 40.0\% & 1.14 \\
Qwen2.5-3B-Instruct (\textbf{Ours, GRPO Cold Start}) & \textbf{60.0\%} & 1.32 \\
\midrule
Qwen2.5-7B-Instruct (\textbf{Raw}) & \uline{69.6\%} & 1.42 \\
Qwen2.5-7B-Instruct (\textbf{SFT400}) & 28.8\% & 3.71 \\
Qwen2.5-7B-Instruct (\textbf{SFT4k}) &  30.4\% & 1.06 \\
Qwen2.5-7B-Instruct (\textbf{SFT400+PPO}) &  45.6\% & 3.54 \\
Qwen2.5-7B-Instruct (\textbf{SFT400+GRPO}) & 29.6\% & 3.70 \\
Qwen2.5-7B-Instruct (\textbf{PPO Cold Start}) & 48.0\% & 1.25 \\
Qwen2.5-7B-Instruct (\textbf{Ours, GRPO Cold Start}) & \textbf{72.0\%} &  1.63\\
\midrule
Llama-3.2-3B-Instruct (\textbf{Raw}) & 34.4\% & 1.25 \\
Llama-3.2-3B-Instruct (\textbf{SFT400})  & 44.0\% & 0.98 \\
Llama-3.2-3B-Instruct (\textbf{SFT4k}) & \uline{48.8\%} & 0.98 \\
Llama-3.2-3B-Instruct (\textbf{SFT400+PPO}) & 39.2\% & 1.33 \\
Llama-3.2-3B-Instruct (\textbf{SFT400+GRPO}) & 45.6\% & 1.00 \\
Llama-3.2-3B-Instruct (\textbf{PPO Cold Start}) & 29.6\% & 1.42 \\
Llama-3.2-3B-Instruct (\textbf{Ours, GRPO Cold Start}) & \textbf{52.0\%} & 0.89 \\
\bottomrule
\end{tabular}
}
\caption{Bamboogle Test Results (Main Result)}
\label{tab:bamboogle-result-main}

\end{minipage}

\end{table*}

\paragraph{Main Results.} We report BFCL and API-Bank results in \Cref{tab:bfcl-result-main} and \Cref{tab:apibank-result-main}, respectively. Our GRPO method, trained from scratch on the Qwen2.5-Instruct series, generally outperforms other baselines, achieving \~10\% absolute gains over SFT trained on the same data volume. In contrast, LLaMA-3.2-Instruct shows less improvement, possibly due to the model's lower adaptability to GRPO-style generalization. Nevertheless, it remains competitive and outperforms most baselines on API-Bank.

\begin{figure}
    \centering
    \subfigure[Format Reward]{
        \includegraphics[width=0.469\linewidth]{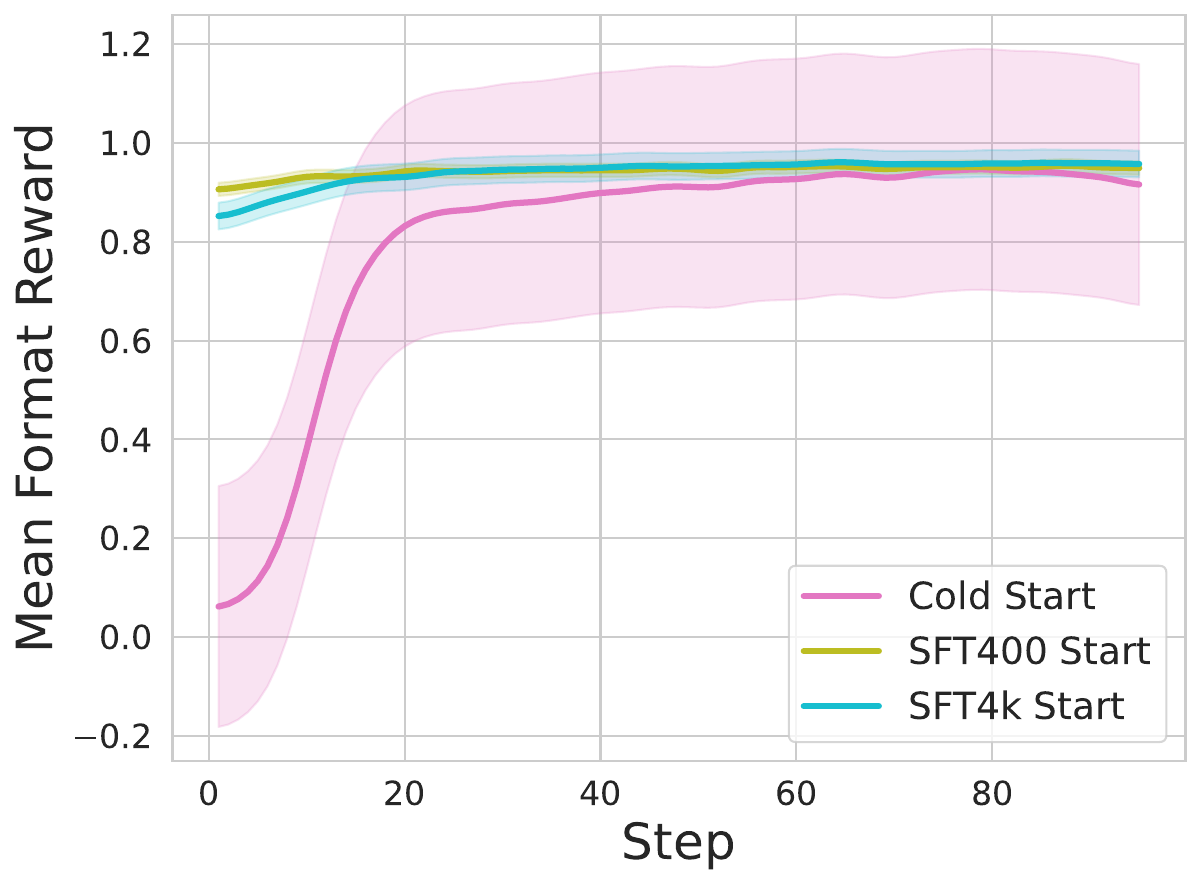}
    }
    \hfill
    \subfigure[Correctness Reward]{
        \includegraphics[width=0.469\linewidth]{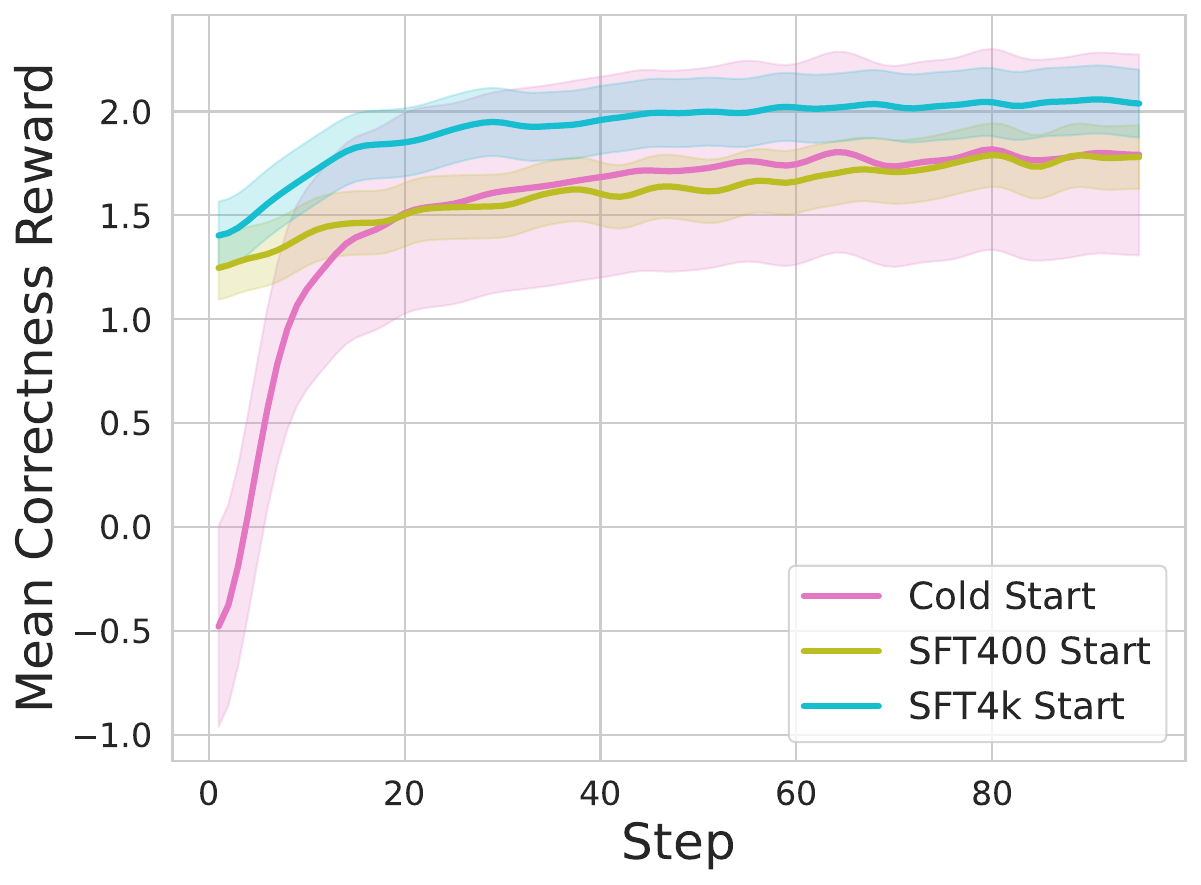}
    }
    \caption{Format (left) and correctness (right) reward trends across training steps for Qwen2.5-3B-Instruct with different model initialization strategies.}
    \label{fig:main_initialization}
\end{figure}

\paragraph{SFT Initialization Impacts.} Interestingly, GRPO also improves models initialized with limited SFT, often outperforming full-scale SFT trained on 10 times more data. However, this setup still underperforms compared to cold start GRPO. We hypothesize that SFT initialization leads to memorization and overfitting, which reduces the impact of GRPO’s effectiveness in generalization. As shown in \Cref{fig:main_initialization}, SFT-initialized models achieve higher training rewards due to distributional alignment between SFT and RL data, but empirically generalize worse on the two benchmarks. This further highlights that higher training rewards do not necessarily translate to better generalization.

\begin{figure}
    \centering
    \subfigure[Format Reward]{
        \includegraphics[width=0.469\linewidth]{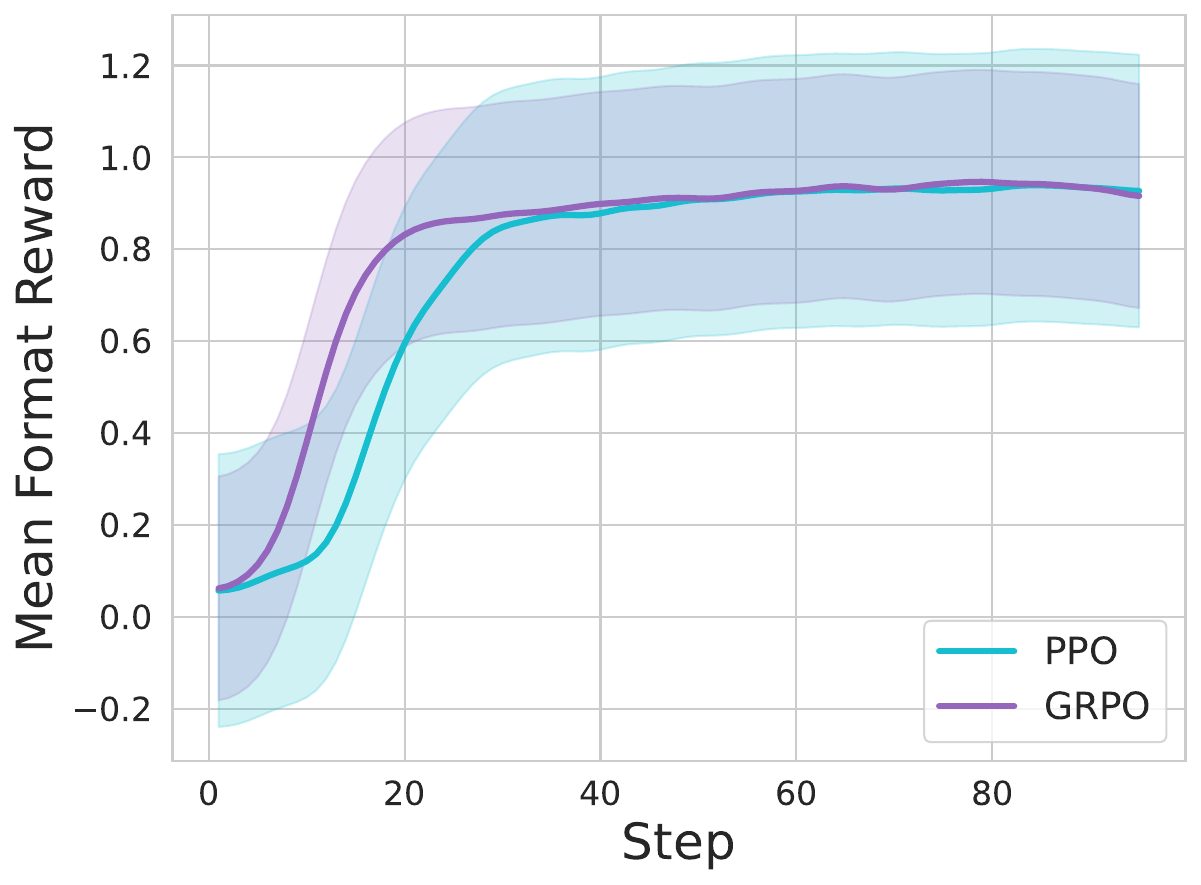}
    }
    \hfill
    \subfigure[Correctness Reward]{
        \includegraphics[width=0.469\linewidth]{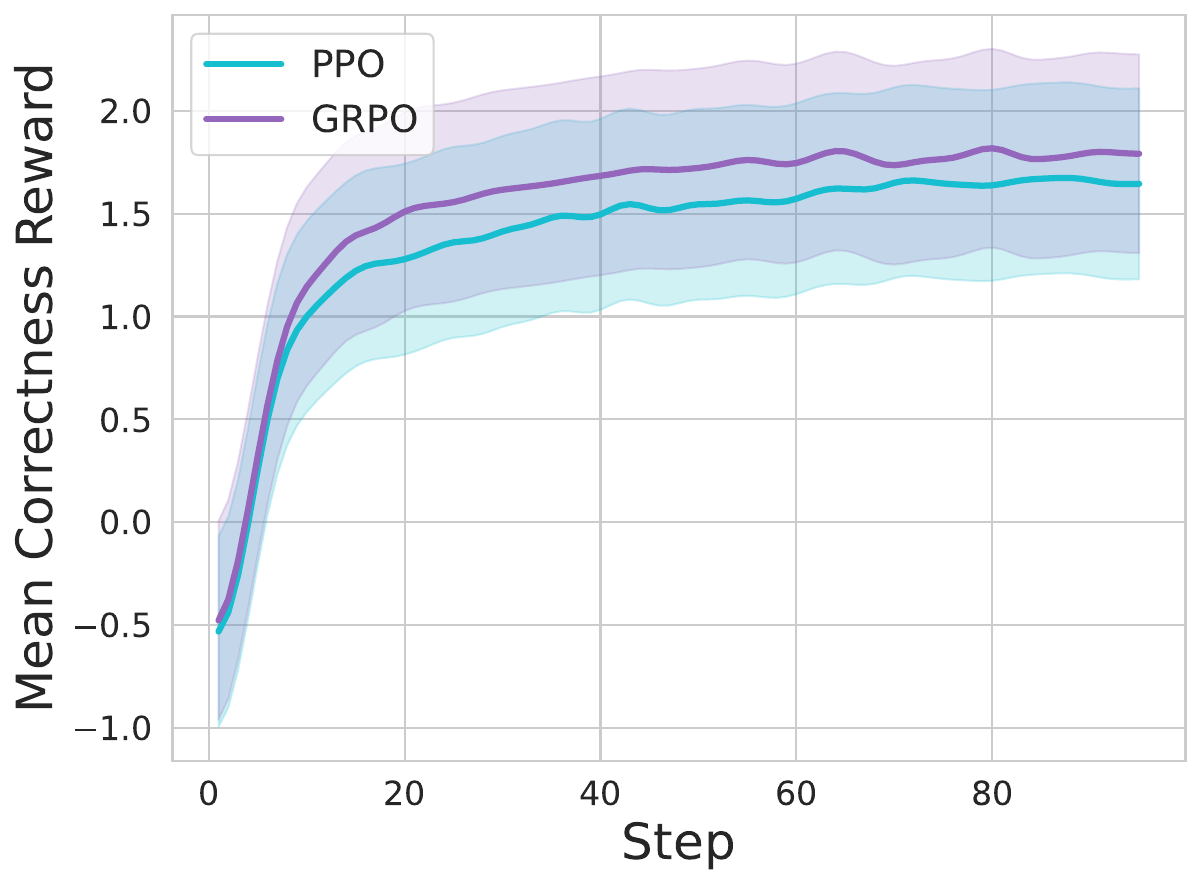}
    }
    \caption{Format (left) and correctness (right) reward trends across training steps for Qwen2.5-3B-Instruct with different RL strategies (GRPO v.s. PPO).}
    \label{fig:main_ppo}
\end{figure}

\paragraph{Reward Design on PPO.} We also evaluate PPO under both cold start and SFT-initialized settings to examine the effectiveness of our reward design. The results show that while PPO with a cold start can outperform SFT in some cases, it tends to be less stable across different model settings. In contrast, GRPO consistently achieves higher rewards even from a cold start, suggesting that our reward design is partially effective for PPO but works best in the GRPO framework. As shown in \Cref{fig:main_ppo}, GRPO not only achieves higher correctness rewards but also gains format rewards more rapidly during training. Interestingly, PPO benefits from SFT initialization, generally yielding better results than a cold start, whereas GRPO performs better when trained from scratch. These findings highlight that while PPO can benefit from our reward design, its impact is more limited compared to the more robust and consistent improvements observed with GRPO.

\begin{figure}
    \centering
    \subfigure[Unfamiliar Scenario]{
        \includegraphics[width=0.469\linewidth]{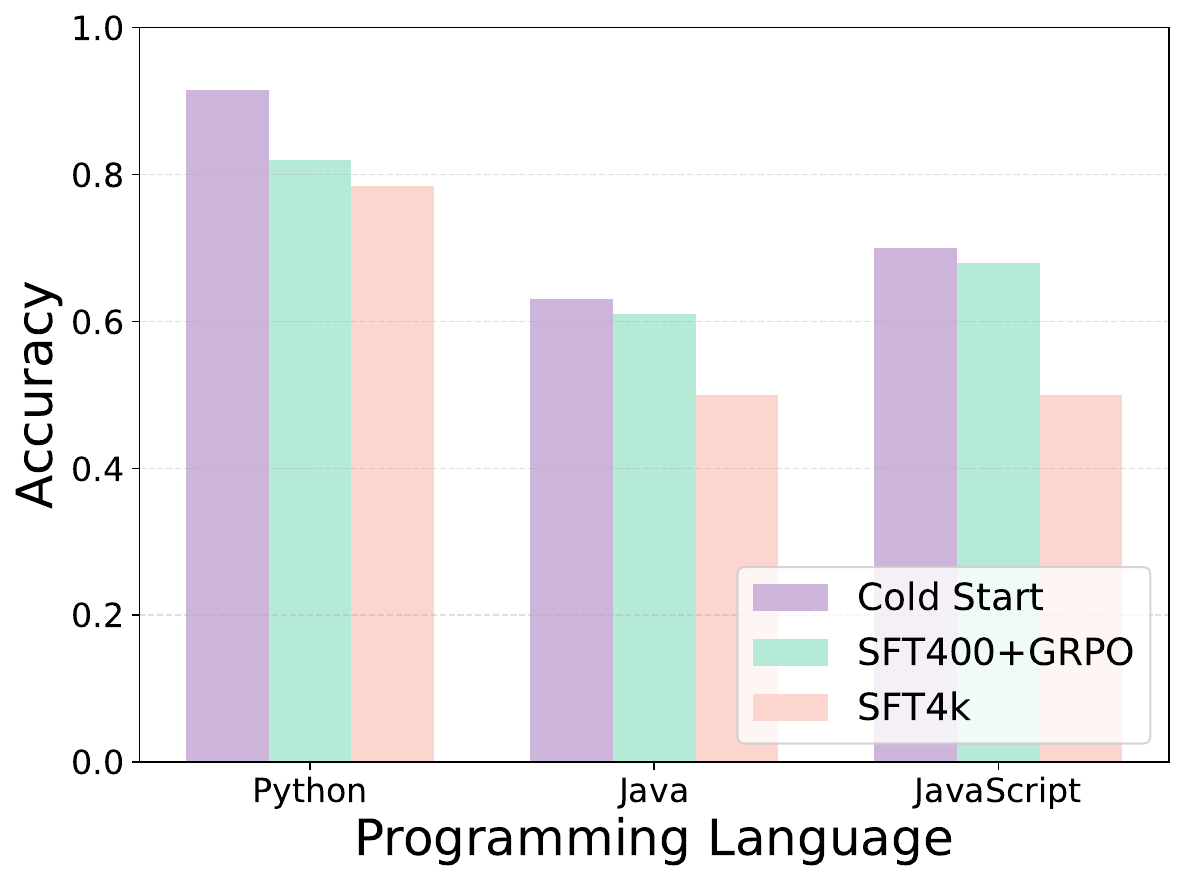}
    }
    \hfill
    \subfigure[Unfamiliar Goal]{
        \includegraphics[width=0.469\linewidth]{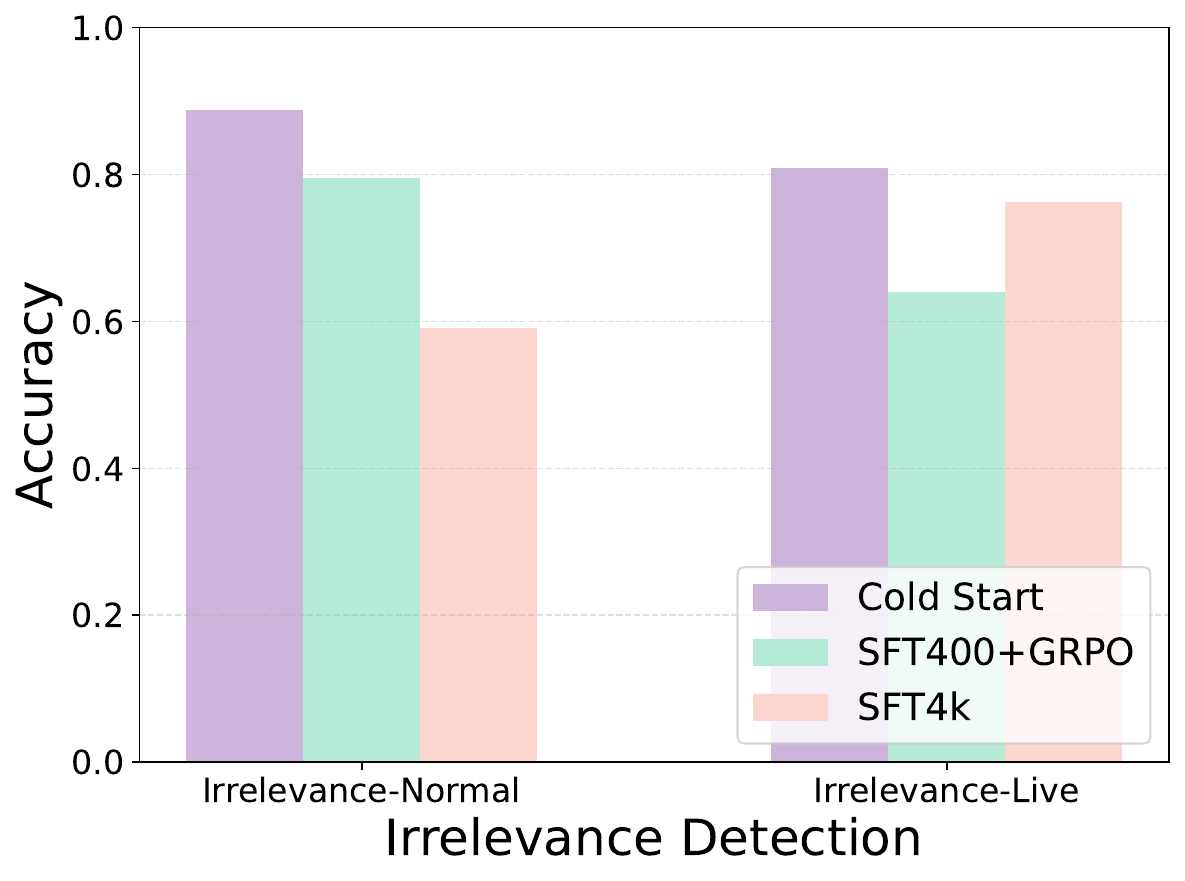}
    }
    \caption{Qwen2.5-3B-Instruct's performance across unfamiliar programming language scenarios (left) and novel relevance detection task goals (right), evaluated under different training settings.}
    \label{fig:main_generalization}
\end{figure}

\begin{table}[t]
\tiny
    \centering
    \begin{minipage}{0.5\textwidth}
        \centering

\begin{tabular}{l}
\toprule 
User: I would like to buy a movie ticket in San Jose at 11 o'clock in the night.\\
Tools Available:\\
\texttt{1. Movies\_1\_BuyMovieTickets(movie\_name, number\_of\_tickets, ...)}  \\
→ Buy tickets for a movie at a specific time and location.\\
\texttt{2. Movies\_1\_FindMovies(location, ...)}  \\
→ Search movies by location, genre, or format.\\
\texttt{3. Movies\_1\_GetTimesForMovie(movie\_name, location, ...)}  \\
→ Get showtimes for a movie on a specific date and location.\\
\midrule
\blue{<think>} The user wants to buy a movie ticket in San Jose at 11 PM. The available\\
tool, \texttt{Movies\_1\_BuyMovieTickets}, allows us to purchase movie tickets with the\\
required details such as movie name, number of tickets, show date, location, show time, \\
and show type. We need to find the movie name and the show date. However, the user \\
\green{did not specify the movie name or the show date}. We can infer the show date from the \\
current date, but we need to know the movie name to proceed with purchase. \blue{</think>} \\
\\
\purple{<response>} To assist in purchasing a movie ticket, \green{I need the name of the movie and} \\
\green{the specific date of the show}. Could you please provide this information? \purple{</response>} \\
\bottomrule
\end{tabular}

    \end{minipage}%
    \vspace{5pt}
    \begin{minipage}{0.5\textwidth}
        \centering

\begin{tabular}{l}
\toprule 
User: What is the slope of the line which is perpendicular to the line with the \\
equation y = 3x + 2?\\
Tools Available:\\
\texttt{1. find\_critical\_points(function, variable, range=[0.0, 3.4])}  \\
→ Find critical points of a function with respect to a variable, optionally \\
within a range.\\
\midrule
\blue{<think>} The problem is asking for the slope of a line that is perpendicular \\
to the line with the equation y = 3x + 2. The slope of the given line is 3. The slope \\
of a line that is perpendicular to another line is the negative reciprocal of the \\
original line's slope. So, the slope of the line perpendicular to y = 3x + 2 would be -1/3.\\
The given tool \texttt{find\_critical\_points} is \green{not relevant to finding the slope of a line and} \\
\green{cannot be used to solve this problem}. \blue{</think>} \\
\\
\purple{<response>} The slope of the line that is perpendicular to the line with the \\
equation y = 3x + 2 is \green{-1/3}. \purple{</response>} \\
\bottomrule
\end{tabular}

    \end{minipage}

    \label{tab:generalization_cases}
    \caption{Cases of Qwen2.5-3B-Instruct (GRPO Cold Start) generalizing to irrelevant tool detection goals.}
    
\end{table}

\paragraph{Generalization Studies.} We evaluate the generalization ability of our trained model in two challenging settings: unfamiliar scenarios and novel task goals (both from BFCL benchmark subset). Specifically, we test the model's performance in tool usage within unseen programming languages and its ability to detect irrelevant tools, neither of which were explicitly included during RL training or in the dataset. As shown in \Cref{fig:main_generalization}, Qwen2.5-3B-Instruct, when trained from scratch with our GRPO-based reward design, consistently achieves highest performance. Additionally, \Cref{tab:generalization_cases} presents two qualitative examples where the model proactively rejects inappropriate tool use—first by clarifying ambiguous intent, and second by opting to answer directly without tools. These behaviors reflect emergent proactivity and metacognition, enhancing efficiency, reducing hallucinations, and signaling foundational agentic intelligence.

\paragraph{Free-form Inference Effectiveness.} While our model is trained with a focus on tool call format and correctness, we further evaluate its ability to handle free-form tool use in a QA setting. Unlike the structured tool selection and application tasks, QA setting: (1) imposes no constraints on tool call parameters, and (2) evaluates only the final answer, making it a ``goal-oriented'' rather than a ``process-oriented'' task. This naturally introduces a multi-step interaction scenario.

Specifically, we use Bamboogle, a multi-hop QA dataset, to assess this capability. The model is equipped with a web search tool, and we report both the answer accuracy and the number of tool calls for all baselines and our approach. As shown in \Cref{tab:bamboogle-result-main}, our reward design achieves the highest performance, despite this setting not being explicitly seen during training. Notably, our cold start GRPO model surpasses others in accuracy without relying on excessive number of tool calls. This suggests that the model can flexibly invoke tools when needed, effectively leverage feedback, wisely and efficiently navigating toward the correct answer.


\section{Analysis}
\label{sec:analysis}
In this section, we conduct a series of ablation studies to identify the most effective reward design for tool calling. We explore various factors including reward type, scale, granularity, and temporal dynamics.

\subsection{Effect of Length Reward}
We first examine the role of a length-based reward. Prior work has demonstrated that the R1-like models can promote deeper reasoning, often reflected in longer thinking traces. To encourage this behavior, we introduce a reward term proportional to the length of the \textcolor{deepblue}{\textless{}think\textgreater{}} field:
\[
\mathcal{R}_\text{length} = \min\left(\frac{L_{\text{think}}}{L_{\text{target}}}, 1\right)
\]
where \( L_{\text{think}} \) denotes the length of the thinking segment in model's output, and \( L_{\text{target}} \) denotes the target output length, which we empirically set to 512. We found that the raw model rarely generates responses longer than half this length, making 512 a reasonable and effective target for encouraging longer outputs. This length-based component is added to the overall reward, which now consists of format, correctness, and reasoning length.

\begin{table*}[!t]
\centering
\resizebox{1.0\linewidth}{!}{
\begin{tabular}{lccccccc}
\toprule
\textbf{Model} & \textbf{Overall Acc} & Non-Live AST Acc & Non-Live Exec Acc & Live Acc & Multi Turn Acc & Relevance Detection & Irrelevance Detection \\
\midrule
Qwen2.5-1.5B-Instruct (\textbf{Original}) & \textbf{46.20\%} & 77.96\% & 76.98\% & 60.73\% & 2.25\% & 100.00\% & 56.44\% \\
Qwen2.5-1.5B-Instruct (\textbf{w/ Length Reward}) & \uline{33.23\%} & 70.58\% & 71.36\% & 35.63\% & 0.50\% & 94.44\% & 4.52\% \\
Qwen2.5-1.5B-Instruct (\textbf{Dynamic}) & 28.51\% & 53.23\% & 48.23\% & 38.07\% & 0.00\% & 55.56\% & 25.08\% \\
\midrule
Qwen2.5-3B-Instruct (\textbf{Original}) & \textbf{52.98\%} & 81.58\% & 79.43\% & 73.78\% & 3.75\% & 88.24\% & 84.85\% \\
Qwen2.5-3B-Instruct (\textbf{w/ Length reward}) & \uline{48.89\%} & 77.83\% & 78.61\% & 63.56\% & 4.50\% & 88.24\% & 71.22\% \\
Qwen2.5-3B-Instruct (\textbf{Dynamic}) & 48.24\% & 77.60\% & 79.11\% & 63.22\% & 3.00\% & 88.89\% & 68.53\% \\
\midrule
Llama-3.2-3B-Instruct (\textbf{Original}) & \uline{44.10\%} & 74.38\% & 75.18\% & 56.86\% & 1.37\% & 94.44\% & 62.23\% \\
Llama-3.2-3B-Instruct (\textbf{w/ Length reward}) & \textbf{44.98\%} & 78.02\% & 77.54\% & 56.55\% & 1.25\% & 100.00\% & 63.76\% \\
Llama-3.2-3B-Instruct (\textbf{Dynamic}) & 43.15\% & 75.50\% & 71.64\% & 56.06\% & 1.00\% & 100.00\% & 57.82\% \\
\bottomrule
\end{tabular}
}
\caption{BFCL V3 Benchmark Results (Length)}
\label{tab:bfcl-result-length}
\end{table*}

\begin{figure}
    \centering
    \subfigure[Response Length]{
        \includegraphics[width=0.469\linewidth]{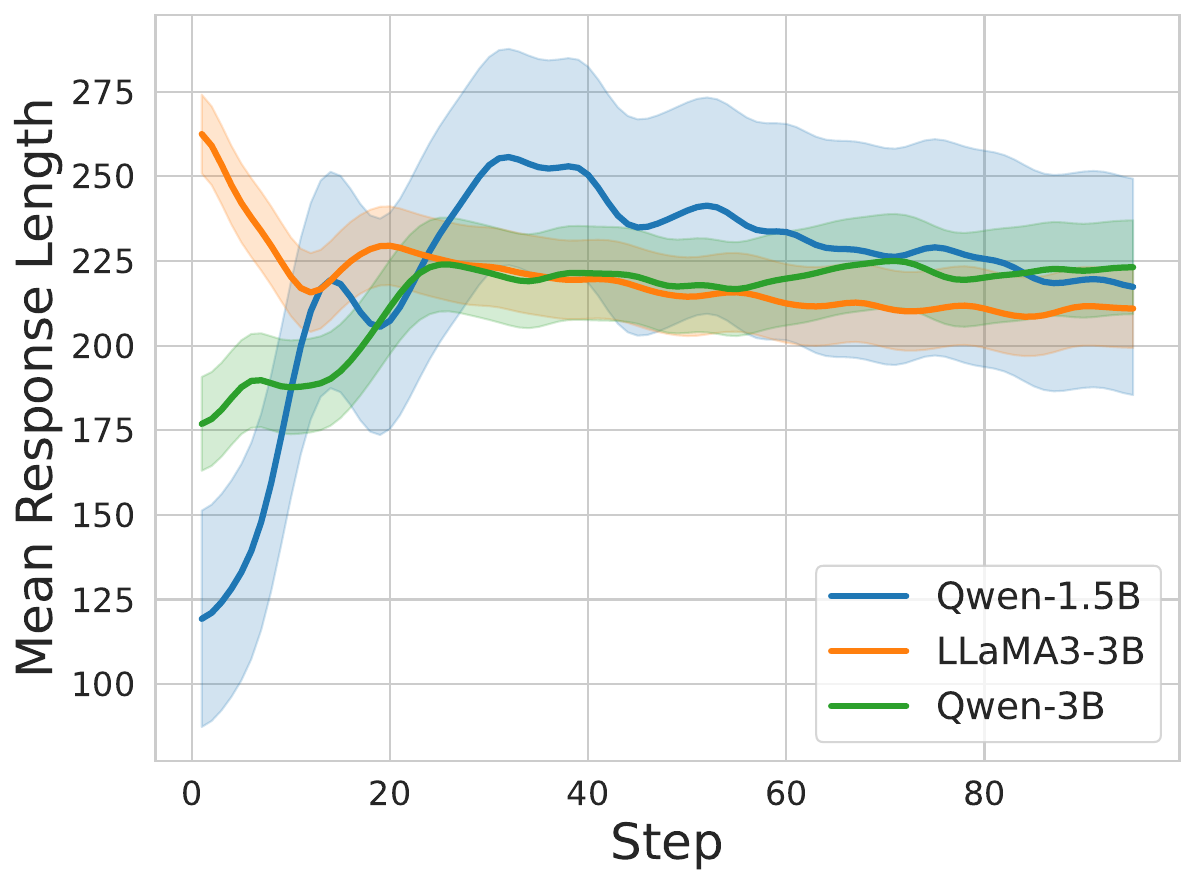}
    }
    \hfill
    \subfigure[Length Reward]{
        \includegraphics[width=0.469\linewidth]{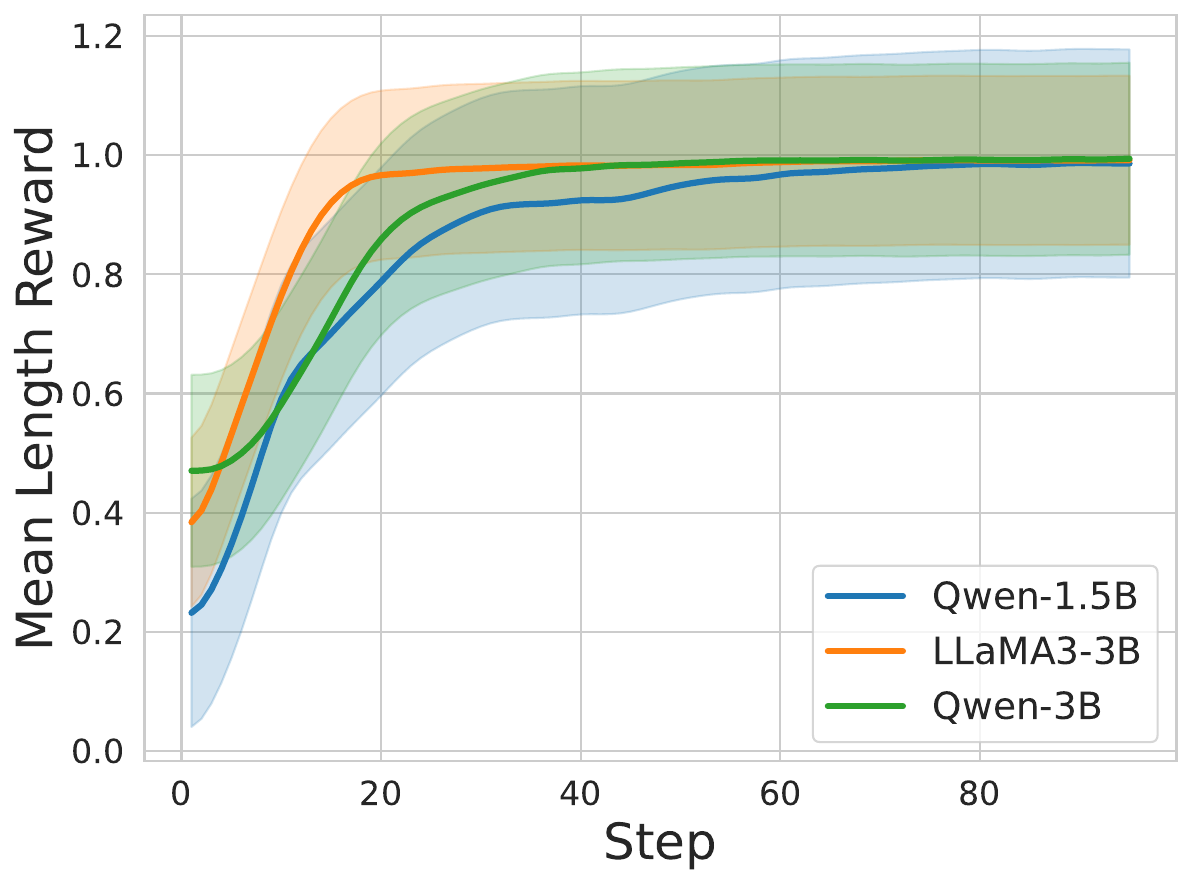}
    }
    \caption{Response length (left) and its reward (right) trends across training steps for different models.}
    \label{fig:analysis_length}
\end{figure}

As shown in \Cref{fig:analysis_length}, both response length and the length reward generally increase throughout training, particularly for the Qwen model series. This indicates that the length reward effectively encourages longer reasoning. However, the downstream results in \Cref{tab:bfcl-result-length} reveal that adding a length reward does not consistently improve task performance, and in smaller-scale models, it can even cause substantial degradation. These observations suggest that while extended reasoning may appear desirable, it is not always beneficial for tool use tasks. In fact, excessive length may introduce unnecessary complexity, leading to overthinking and reduced effectiveness.

\paragraph{Dynamic Length Reward.} Since fixed-length rewards showed minimal impact and converged quickly, we explored a dynamic length reward that adapts over training steps. Specifically, we define:
\[
\mathcal{R}_\text{dynamic} = \min\left(\frac{L_{\text{think}}}{L_{\text{target}} \cdot (1 + p) }, 1\right)
\]
where \( S \) denotes the training steps and \( p = \frac{S_{\text{current}}}{S_{\text{total}}} \in [0, 1] \) represents the normalized training progress. This formulation gradually increases the target thinking length over time, aligning with model maturity.

\begin{figure}
    \centering
    \subfigure[Response Length]{
        \includegraphics[width=0.469\linewidth]{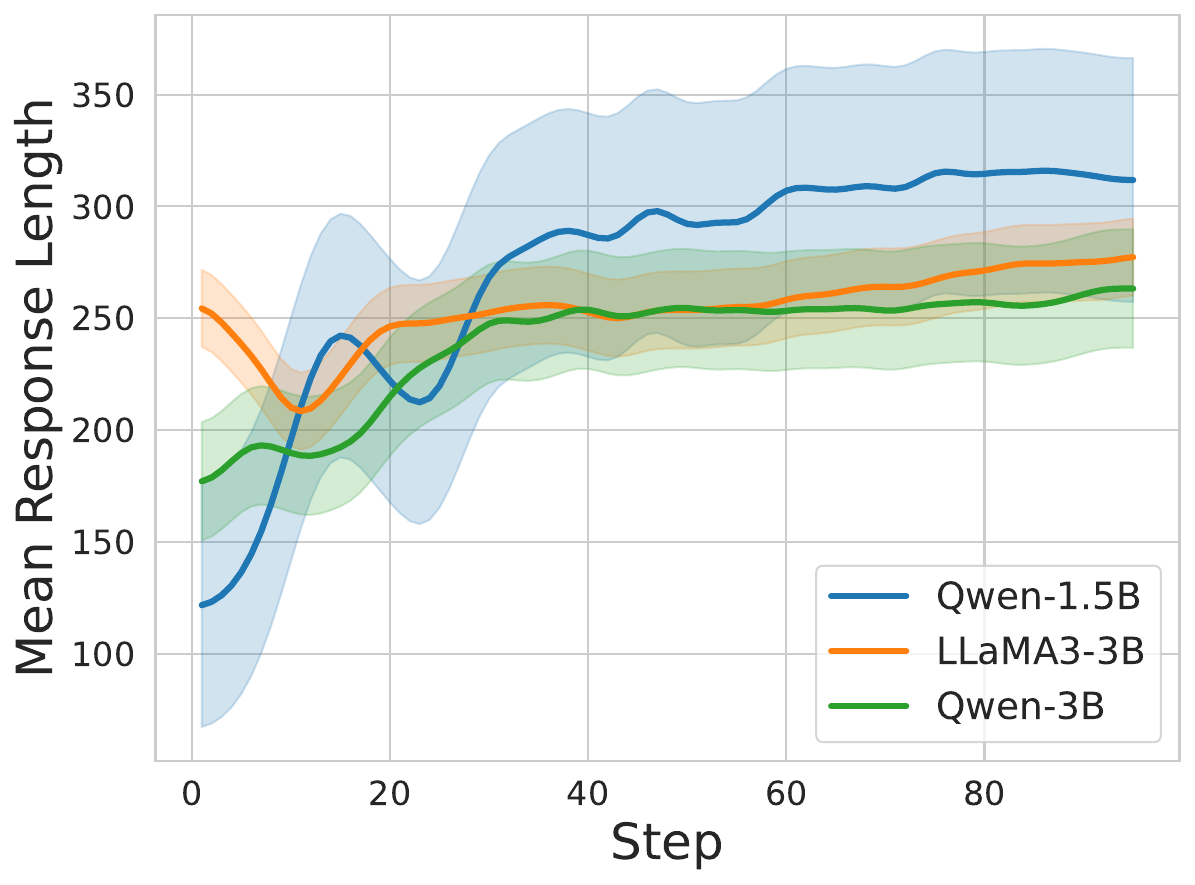}
    }
    \hfill
    \subfigure[Length Reward]{
        \includegraphics[width=0.469\linewidth]{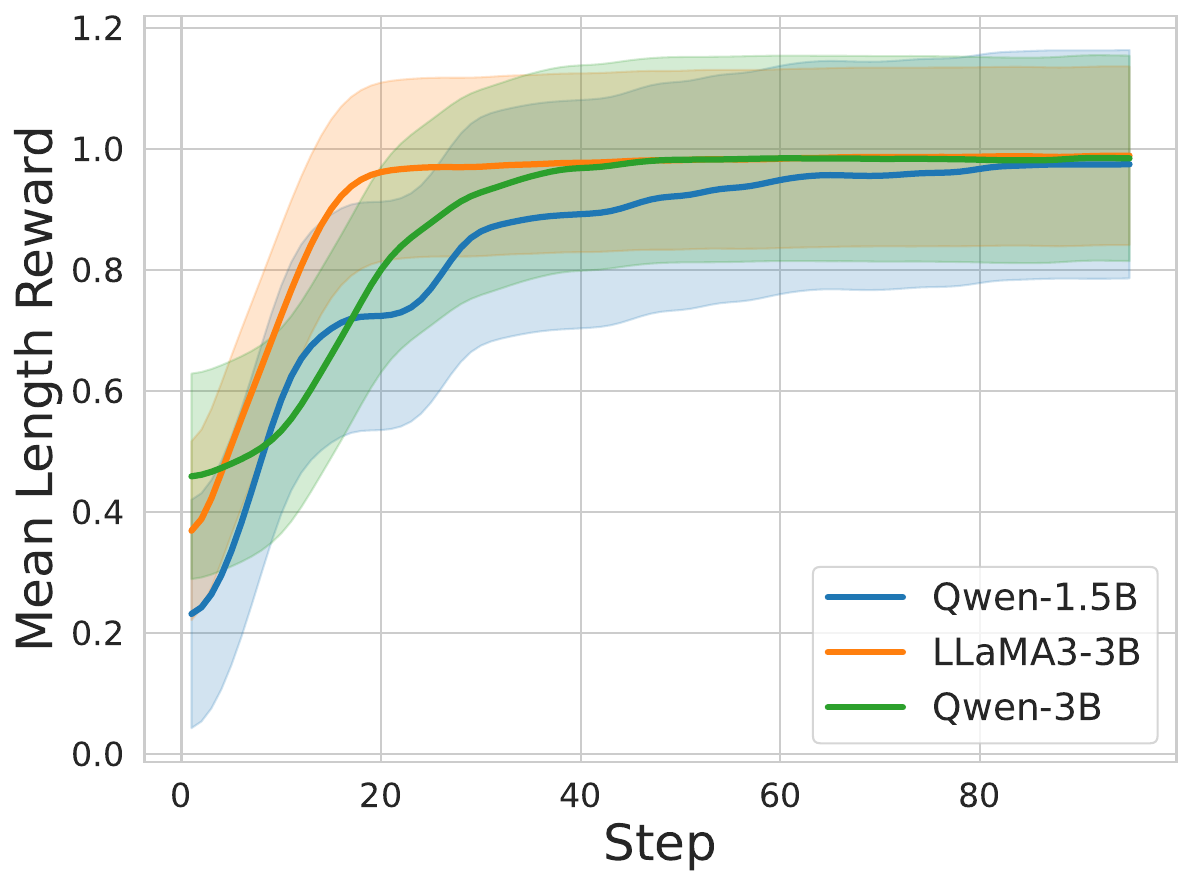}
    }
    \caption{Response length (left) and its reward (right) trends across training steps within the dynamic length reward training setting.}
    \label{fig:analysis_length_dynamic}
\end{figure}

As shown in \cref{fig:analysis_length_dynamic}, this approach yields a steadier growth in thinking length, particularly for the Llama model. However, the performance results in \Cref{tab:bfcl-result-length} reveal that even scheduled rewards fail to improve performance. This further supports our hypothesis that extended reasoning may not benefit this task and can even have adverse effects.

\finding{\textbf{Takeaway 1}: While length rewards encourage longer reasoning traces, they do not consistently improve task performance and may even harm it in smaller models, highlighting that longer reasoning is not inherently better for tool use tasks.}

\subsection{Effect of Reward Scale}
Next, we investigate the effect of reward scaling, specifically the relative weighting between correctness and format rewards. Prior work in R1-style RL commonly assigns a higher weight to correctness reward than to format reward~\citep{xie2025logic, jin2025search}, emphasizing the importance of learning correct answer over superficial adherence to format. This strategy helps prevent reward hacking, where a model might exploit formatting heuristics without learning task semantics.

To test the importance of this design choice, we conduct an ablation where we equalize the maximum correctness and format rewards by setting the former’s range to \([-1, 1]\), matching that of the format reward. This adjustment only affects the final normalization step of the correctness reward:
\begin{equation*}
\mathcal{R}_{\text{correct}} = 2 \cdot \frac{R_{\max}}{S_{\max}} - 1 \in [-1, 1]
\end{equation*}
where all variables are defined as in \Cref{sec:reward_design}.

\begin{table*}[!t]
\centering
\resizebox{1.0\linewidth}{!}{
\begin{tabular}{lccccccc}
\toprule
\textbf{Model} & \textbf{Overall Acc} & Non-Live AST Acc & Non-Live Exec Acc & Live Acc & Multi Turn Acc & Relevance Detection & Irrelevance Detection \\
\midrule
Qwen2.5-1.5B-Instruct (\textbf{Original}) & \textbf{46.20\%} & 77.96\% & 76.98\% & 60.73\% & 2.25\% & 100.00\% & 56.44\% \\
Qwen2.5-1.5B-Instruct (\textbf{Equal max}) & 39.47\% & 78.56\% & 75.50\% & 45.45\% & 2.50\% & 100.00\% & 16.44\% \\
Qwen2.5-1.5B-Instruct (\textbf{Two stage}) & 38.85\% & 77.96\% & 76.23\% & 44.51\% & 2.25\% & 100.00\% & 10.61\% \\
Qwen2.5-1.5B-Instruct (\textbf{Dynamic}) & \uline{45.71\%} & 78.31\% & 75.73\% & 58.91\% & 2.50\% & 100.00\% & 57.20\% \\
\midrule
Qwen2.5-3B-Instruct (\textbf{Original}) & \uline{52.98\%} & 81.58\% & 79.43\% & 73.78\% & 3.75\% & 88.24\% & 84.85\% \\
Qwen2.5-3B-Instruct (\textbf{Equal max}) & 51.76\% & 81.50\% & 79.50\% & 69.79\% & 4.25\% & 88.89\% & 78.07\% \\
Qwen2.5-3B-Instruct (\textbf{Two stage}) & 50.66\% & 80.62\% & 78.82\% & 67.93\% & 3.50\% & 88.89\% & 76.42\% \\
Qwen2.5-3B-Instruct (\textbf{Dynamic}) & \textbf{53.81\%} & 81.44\% & 80.75\% & 75.43\% & 3.62\% & 77.78\% & 88.82\% \\
\midrule
Llama-3.2-3B-Instruct (\textbf{Original}) & \uline{44.10\%} & 74.38\% & 75.18\% & 56.86\% & 1.37\% & 94.44\% & 62.23\% \\
Llama-3.2-3B-Instruct (\textbf{Equal max}) & 42.47\% & 67.77\% & 75.05\% & 55.75\% & 1.00\% & 88.89\% & 59.56\% \\
Llama-3.2-3B-Instruct (\textbf{Two stage}) & 41.33\% & 65.54\% & 72.70\% & 55.22\% & 0.75\% & 88.89\% & 57.59\% \\
Llama-3.2-3B-Instruct (\textbf{Dynamic}) & \textbf{46.85\%} & 83.00\% & 72.77\% & 61.00\% & 3.38\% & 88.89\% & 59.37\% \\
\bottomrule
\end{tabular}
}
\caption{BFCL V3 Benchmark Results (Scale)}
\label{tab:bfcl-result-scale}
\end{table*}

As shown in \Cref{tab:bfcl-result-scale}, this equal-scaling variant, denoted as ``Equal Max'', results in a slight drop in overall accuracy across most models, with the exception of Qwen2.5-3B, which maintains performance comparable to the original setting. These results underscore the importance of assigning greater weight to correctness reward: doing so helps steer the model toward mastering the core reasoning and tool use capabilities necessary for robust generalization.

\paragraph{Dynamic Reward Scaling.} Building on the insight that correctness reward plays a more critical role, we are further motivated by the intuition that different reward components may benefit from being emphasized at different stages of training. This leads us to explore dynamically adjusting reward scales in accordance with training progress. Specifically, we hypothesize that in early training, the model should prioritize learning the correct output format, which entails an easier objective, before gradually shifting focus to the more challenging goal of tool use correctness. To test this hypothesis, we design two dynamic reward scaling strategies:

\begin{itemize}[topsep=2pt, leftmargin=10pt, itemsep=2pt]
    \item \textbf{Two stage (Coarse) Setting}: We divide training into two phases. In the first \( s \) training steps, we downscale the correctness reward to \( \frac{1}{3}\) of its original scale while keeping the format reward at its original scale. After step \( s \), we restore the correctness reward to its original scale and simultaneously reduce the format reward to range \([0, 0.5]\) (\( \frac{1}{2}\) of its original scale). Formally the reward scales are:
    \[
    \text{Scale}_\text{format} = 
    \begin{cases}
        [0, 1] & \text{if } S_{\text{current}} < s \\
        [0, 0.5] & \text{otherwise}
    \end{cases},
    \]
    \[
    \text{Scale}_\text{correct} = 
    \begin{cases}
        [-1, 1] & \text{if } S_{\text{current}} < s \\
        [-3, 3] & \text{otherwise}
    \end{cases}
    \]
    where \( S_{\text{current}} \) denotes the current training step. In our experiments, we empirically set the switching point to \( s = 30 \) steps, as we observed that the format reward typically experiences a significant increase within the first 30 steps. Therefore, it is more beneficial for later steps to shift focus toward optimizing correctness.

    \item \textbf{Dynamic (Finegrained) Setting}: We apply continuous interpolation between the two reward scales throughout training. Initially, both the format and correctness reward scales are set equally. Over time, the format reward scale linearly decays to its original value, while the correctness reward scale gradually increases to its original value, allowing the training to shift focus from format adherence to task correctness accordingly. Formally, the dynamic scaling is then defined as:
    \[
    \text{Scale}_\text{format} = [-2 + p, 2 - p],
    \]
    \[
    \text{Scale}_\text{correct} = [-2 - p, 2 + p]
    \]
    where \( p \in [0,1] \) similarly represents the normalized training progress. This design ensures a smooth shift of learning focus from format fidelity to correctness.
\end{itemize}

\begin{figure}
    \centering
    \subfigure[Format Reward]{
        \includegraphics[width=0.469\linewidth]{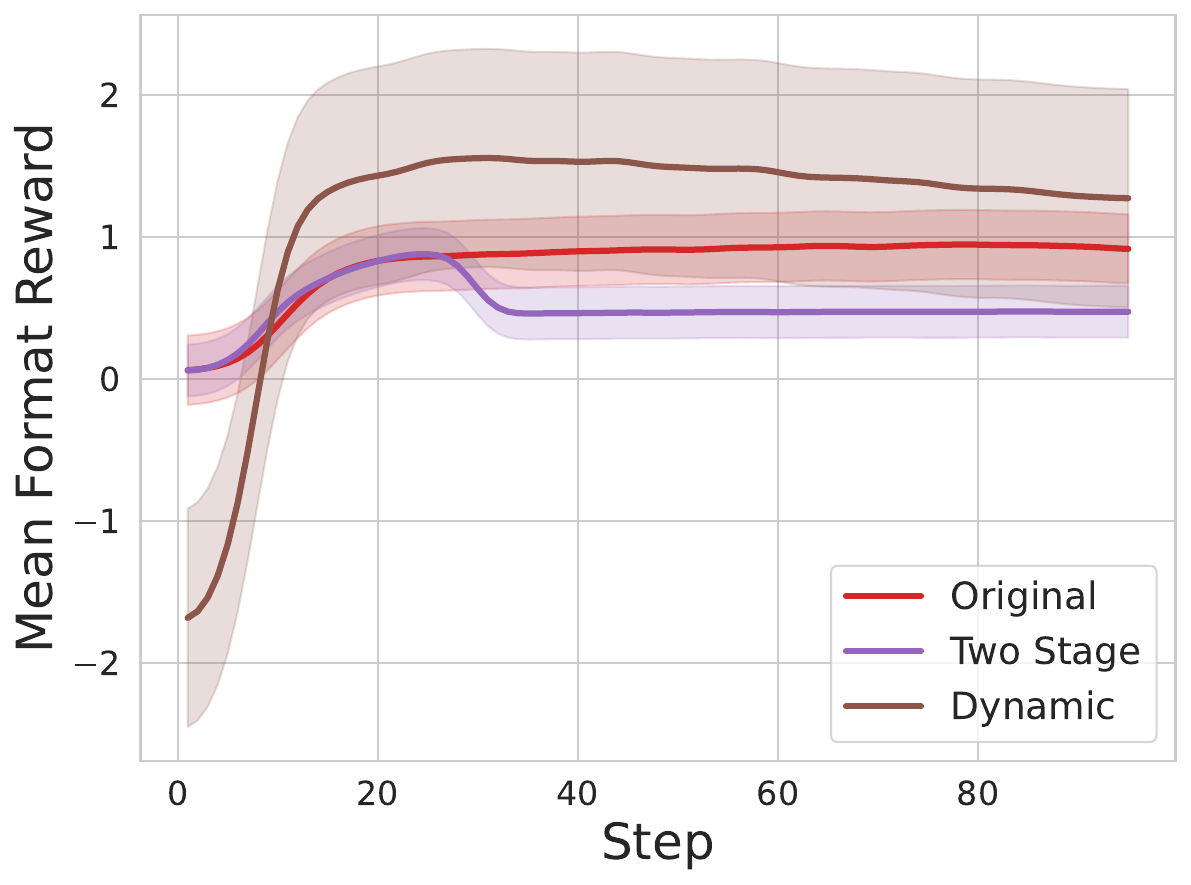}
    }
    \hfill
    \subfigure[Correctness Reward]{
        \includegraphics[width=0.469\linewidth]{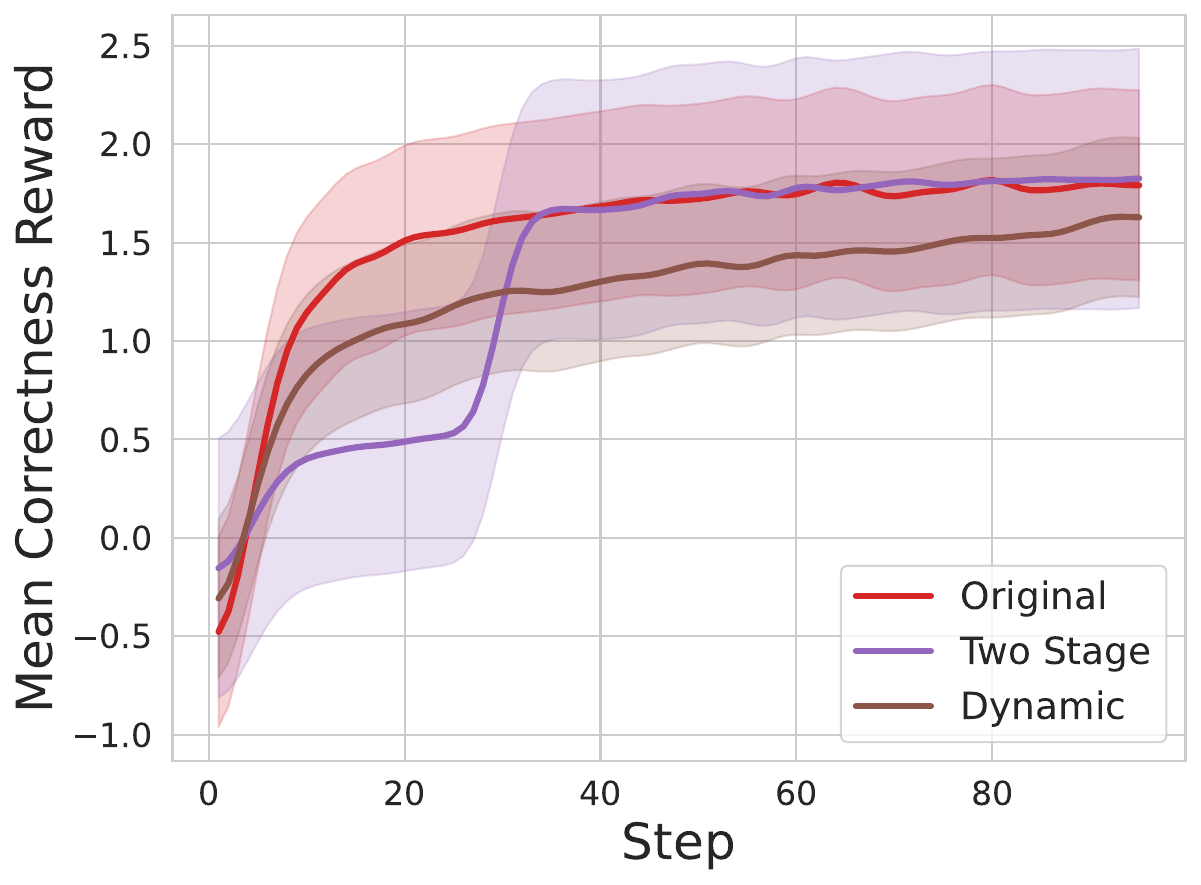}
    }
    \caption{Format (left) and correctness (right) reward trends across training steps for Qwen2.5-3B-Instruct with different reward scale dynamics.}
    \label{fig:analysis_scale_dynamic}
\end{figure}

We present the reward dynamics of the original and two dynamic scaling strategies in \Cref{fig:analysis_scale_dynamic}. As shown in \Cref{tab:bfcl-result-scale}, the Two stage (Coarse) reward setting unexpectedly leads to a drop in performance, whereas the Dynamic (Finegrained) scaling could improve model's benchmarking performance. These findings suggest that abrupt shifts in reward scale may negatively impact the training dynamics. In contrast, a smoother and gradual transition from simpler objectives to more nuanced ones appears to better support the model’s learning trajectory and generalization during GRPO training.

\finding{\textbf{Takeaway 2}: Gradually adjusting reward scales during training, rather than abrupt changes, better supports model learning and generalization, highlighting the benefits of a smoother transition from simpler objectives to more complex ones.}

\subsection{Effect of Reward Granularity}
We now perform a detailed analysis of the effect of reward granularity, focusing specifically on the correctness reward. Tool calling, by nature, poses challenges for reward assignment, as it involves multiple facets beyond a single definitive answer (e.g., in contrast to math reasoning tasks). Our original reward design decomposes correctness into matching the tool name, parameter names, and parameter values, offering a finegrained, ``process-oriented'' signal that reflects partial correctness in tool usage.

To assess the impact of this granularity, we evaluate three alternative reward formulations with progressively coarser levels of aggregation:
\begin{itemize}[topsep=2pt, leftmargin=10pt, itemsep=2pt]

    \item \textbf{Finegrained}: We apply strict exact-match constraints to both tool name and parameter name matching. Specifically, we define:
    
    \begin{small}
    \begin{equation*}
        r_{\text{name}} = \mathds{1}[N_G = N_P] \in \{0, 1\}
    \end{equation*}
    \begin{equation*}
        r_{\text{param}} = \sum_{G_j \in G} \mathds{1}[\text{keys}(P_G) = \text{keys}(P_P)] \in [0, |G|]
    \end{equation*}
    \end{small}
    
    \item \textbf{Intermediate}: We combine the parameter name and value rewards into a single term that enforces an exact match on the entire parameter dictionary. Formally:
    
    \begin{small}
    \begin{equation*}
        r_{\text{param}} + r_{\text{value}} = \sum_{G_j \in G} \mathds{1}[P_G = P_P] \in [0, |G|]
    \end{equation*}
    \end{small}
    
    \item \textbf{Coarse}: At the coarsest level, we fully entangle tool name, parameter names, and parameter values, treating the entire tool set as a unit. Reward is given only if the generated tool set exactly matches the ground truth:
    
    \begin{small}
    \begin{equation*}
        r_{\text{name}} + r_{\text{param}} + r_{\text{value}} = \mathds{1}[G = P] \in \{0, 1\}
    \end{equation*}
    \end{small}

\end{itemize}
All other aspects of reward computation are kept identical to those described in \Cref{sec:reward_design}. Starting from our original design, which is the most finegrained, we progressively entangle reward components to derive increasingly coarse-grained alternatives. 

\begin{figure}
    \centering
    \includegraphics[width=\linewidth]{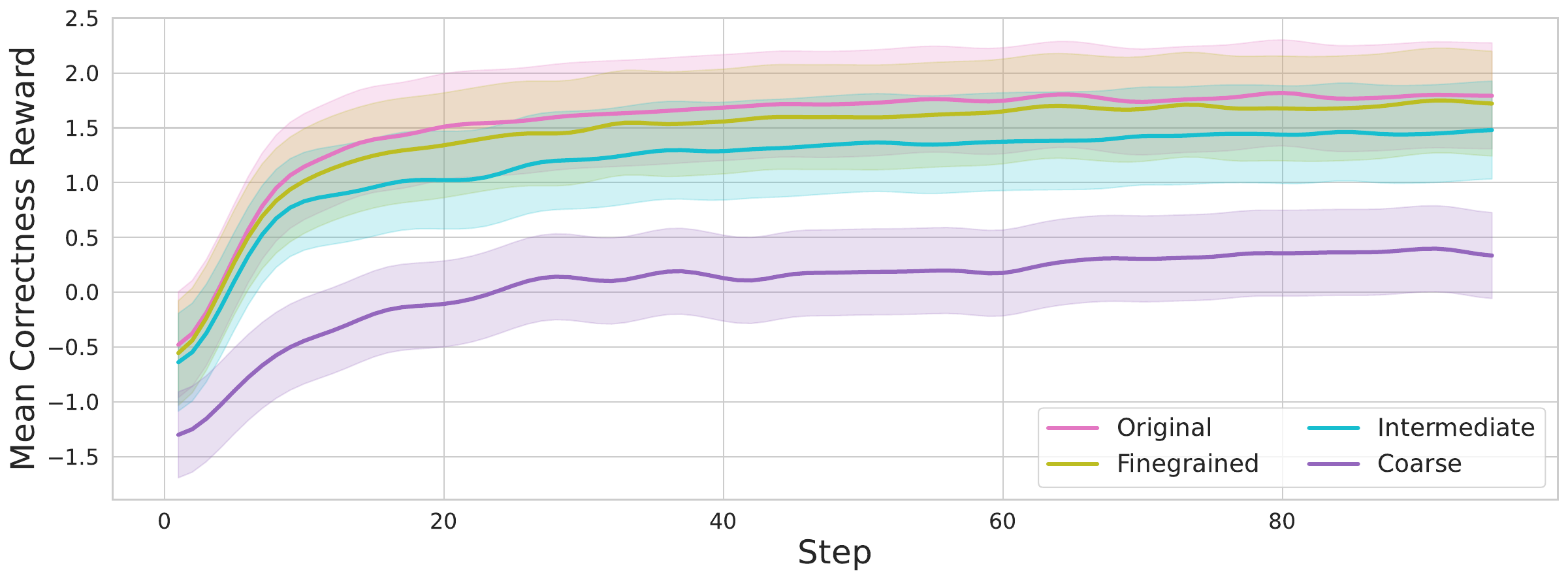}
    \caption{Correctness reward trends across training steps for Qwen2.5-3B-Instruct with different reward granularity.}
    \label{fig:analysis_granularity}
\end{figure}

\begin{table*}[!t]
\centering
\resizebox{1.0\linewidth}{!}{
\begin{tabular}{lccccccc}
\toprule
\textbf{Model} & \textbf{Overall Acc} & Non-Live AST Acc & Non-Live Exec Acc & Live Acc & Multi Turn Acc & Relevance Detection & Irrelevance Detection \\
\midrule
Qwen2.5-1.5B-Instruct (\textbf{Original}) & \textbf{46.20\%} & 77.96\% & 76.98\% & 60.73\% & 2.25\% & 100.00\% & 56.44\% \\
Qwen2.5-1.5B-Instruct (\textbf{Finegrained}) & \uline{40.71\%} & 78.00\% & 75.55\% & 48.91\% & 2.00\% & 100.00\% & 24.84\% \\
Qwen2.5-1.5B-Instruct (\textbf{Intermediate}) & 37.65\% & 77.94\% & 72.46\% & 43.00\% & 1.62\% & 100.00\% & 12.45\% \\
Qwen2.5-1.5B-Instruct (\textbf{Coarse}) & 36.72\% & 76.44\% & 70.86\% & 41.27\% & 2.12\% & 100.00\% & 12.24\% \\
\midrule
Qwen2.5-3B-Instruct (\textbf{Original}) & \textbf{52.98\%} & 81.58\% & 79.43\% & 73.78\% & 3.75\% & 88.24\% & 84.85\% \\
Qwen2.5-3B-Instruct (\textbf{Finegrained}) & \uline{52.06\%} & 81.65\% & 79.64\% & 69.21\% & 5.50\% & 83.33\% & 78.14\% \\
Qwen2.5-3B-Instruct (\textbf{Intermediate}) & 51.36\% & 81.15\% & 80.07\% & 68.64\% & 4.25\% & 88.89\% & 75.74\% \\
Qwen2.5-3B-Instruct (\textbf{Coarse}) & 51.40\% & 79.48\% & 78.54\% & 68.73\% & 5.62\% & 88.89\% & 77.80\% \\
\midrule
Llama-3.2-3B-Instruct (\textbf{Original}) & \textbf{44.10\%} & 74.38\% & 75.18\% & 56.86\% & 1.37\% & 94.44\% & 62.23\% \\
Llama-3.2-3B-Instruct (\textbf{Finegrained}) & \uline{39.82\%} & 64.71\% & 70.68\% & 52.20\% & 0.25\% & 100.00\% & 56.68\% \\
Llama-3.2-3B-Instruct (\textbf{Intermediate}) & 38.62\% & 59.83\% & 71.86\% & 50.56\% & 0.25\% & 94.44\% & 55.68\% \\
Llama-3.2-3B-Instruct (\textbf{Coarse}) & 35.95\% & 52.00\% & 61.43\% & 48.96\% & 1.12\% & 83.33\% & 61.92\% \\
\bottomrule
\end{tabular}
}
\caption{BFCL V3 Benchmark Results (Granularity)}
\label{tab:bfcl-result-granularity}
\end{table*}


The reward dynamics across training steps, shown in \Cref{fig:analysis_granularity}, demonstrate that as reward granularity becomes coarser, it becomes harder for the model to achieve higher reward values during RL training. This suggests that overly strict and entangled rewards may lead to sparse learning signals, potentially hindering effective credit assignment.

Empirical results in \Cref{tab:bfcl-result-granularity} further support this insight: our original, most finegrained reward strategy performs well across models. In general, finer-grained reward decomposition leads to better training outcomes and higher final task performance, indicating its advantage in promoting more stable and effective policy learning.

\finding{\textbf{Takeaway 3}: Finegrained reward decomposition provides richer learning signals, highlighting its role in enabling more effective training compared to coarse reward formulations, which can impede progress and degrade final performance.}

\section{Conclusion}
In this paper, we present a reward design tailored for GRPO training on tool use tasks. Empirically, our model trained from scratch using GRPO consistently outperforms both SFT-based and SFT-initialized RL baselines, as well as models trained with alternative RL algorithms, across a variety of held-out tool use benchmarks. Furthermore, we demonstrate that our model generalizes well to QA settings, exhibiting robust multi-turn interactions, emergent proactiveness, and metacognitive behaviors, all of which are key traits for efficient and adaptable tool use, lying at the core of foundational agent capabilities.
Our in-depth analysis of reward types, scaling strategies, granularity, and temporal dynamics provides further insights into how reward shaping influences learning and behavior. We hope these findings serve as a roadmap for future work in applying reinforcement learning to tool use. Ultimately, we envision that \textit{reward is all tool learning needs}, and that RL offers a powerful path toward generalizable and creative agent behavior.


\bibliography{custom}

\begin{thebibliography}{60}
\providecommand{\natexlab}[1]{#1}

\bibitem[{Acikgoz et~al.(2025)Acikgoz, Greer, Datta, Yang, Zeng, Elachqar, Koukoumidis, Hakkani-Tür, and Tur}]{acikgoz2025coalm}
Emre~Can Acikgoz, Jeremiah Greer, Akul Datta, Ze~Yang, William Zeng, Oussama Elachqar, Emmanouil Koukoumidis, Dilek Hakkani-Tür, and Gokhan Tur. 2025.
\newblock \href {https://arxiv.org/abs/2502.08820} {Can a single model master both multi-turn conversations and tool use? coalm: A unified conversational agentic language model}.
\newblock \emph{Preprint}, arXiv:2502.08820.

\bibitem[{Baek et~al.(2024)Baek, Jauhar, Cucerzan, and Hwang}]{baek2024researchagent}
Jinheon Baek, Sujay~Kumar Jauhar, Silviu Cucerzan, and Sung~Ju Hwang. 2024.
\newblock Researchagent: Iterative research idea generation over scientific literature with large language models.
\newblock \emph{arXiv preprint arXiv:2404.07738}.

\bibitem[{Chen et~al.(2023{\natexlab{a}})Chen, Shu, Shareghi, Collier, Narasimhan, and Yao}]{chen2023fireact}
Baian Chen, Chang Shu, Ehsan Shareghi, Nigel Collier, Karthik Narasimhan, and Shunyu Yao. 2023{\natexlab{a}}.
\newblock Fireact: Toward language agent fine-tuning.
\newblock \emph{arXiv preprint arXiv:2310.05915}.

\bibitem[{Chen et~al.(2023{\natexlab{b}})Chen, Li, Wang, and Li}]{chen2023good}
Nuo Chen, Hongguang Li, Baoyuan Wang, and Jia Li. 2023{\natexlab{b}}.
\newblock From good to great: Improving math reasoning with tool-augmented interleaf prompting.
\newblock \emph{arXiv preprint arXiv:2401.05384}.

\bibitem[{Chen et~al.(2022)Chen, Ma, Wang, and Cohen}]{chen2022program}
Wenhu Chen, Xueguang Ma, Xinyi Wang, and William~W Cohen. 2022.
\newblock Program of thoughts prompting: Disentangling computation from reasoning for numerical reasoning tasks.
\newblock \emph{arXiv preprint arXiv:2211.12588}.

\bibitem[{Chen et~al.(2024)Chen, Liu, Wang, Zhang, Liu, Lin, Chen, and Zhao}]{chen2024agentflan}
Zehui Chen, Kuikun Liu, Qiuchen Wang, Wenwei Zhang, Jiangning Liu, Dahua Lin, Kai Chen, and Feng Zhao. 2024.
\newblock \href {https://doi.org/10.18653/v1/2024.findings-acl.557} {Agent-{FLAN}: Designing data and methods of effective agent tuning for large language models}.
\newblock In \emph{Findings of the Association for Computational Linguistics: ACL 2024}, pages 9354--9366, Bangkok, Thailand. Association for Computational Linguistics.

\bibitem[{Chu et~al.(2025)Chu, Zhai, Yang, Tong, Xie, Schuurmans, Le, Levine, and Ma}]{chu2025sft}
Tianzhe Chu, Yuexiang Zhai, Jihan Yang, Shengbang Tong, Saining Xie, Dale Schuurmans, Quoc~V Le, Sergey Levine, and Yi~Ma. 2025.
\newblock Sft memorizes, rl generalizes: A comparative study of foundation model post-training.
\newblock \emph{arXiv preprint arXiv:2501.17161}.

\bibitem[{Dang and Ngo(2025)}]{dang2025reinforcement}
Quy-Anh Dang and Chris Ngo. 2025.
\newblock Reinforcement learning for reasoning in small llms: What works and what doesn't.
\newblock \emph{arXiv preprint arXiv:2503.16219}.

\bibitem[{Dubey et~al.(2024)Dubey, Jauhri, Pandey, Kadian, Al-Dahle, Letman, Mathur, Schelten, Yang, Fan et~al.}]{dubey2024llama}
Abhimanyu Dubey, Abhinav Jauhri, Abhinav Pandey, Abhishek Kadian, Ahmad Al-Dahle, Aiesha Letman, Akhil Mathur, Alan Schelten, Amy Yang, Angela Fan, et~al. 2024.
\newblock The llama 3 herd of models.
\newblock \emph{arXiv preprint arXiv:2407.21783}.

\bibitem[{Gou et~al.(2023)Gou, Shao, Gong, Shen, Yang, Huang, Duan, and Chen}]{gou2023tora}
Zhibin Gou, Zhihong Shao, Yeyun Gong, Yelong Shen, Yujiu Yang, Minlie Huang, Nan Duan, and Weizhu Chen. 2023.
\newblock Tora: A tool-integrated reasoning agent for mathematical problem solving.
\newblock \emph{arXiv preprint arXiv:2309.17452}.

\bibitem[{Guo et~al.(2025)Guo, Yang, Zhang, Song, Zhang, Xu, Zhu, Ma, Wang, Bi et~al.}]{guo2025deepseek}
Daya Guo, Dejian Yang, Haowei Zhang, Junxiao Song, Ruoyu Zhang, Runxin Xu, Qihao Zhu, Shirong Ma, Peiyi Wang, Xiao Bi, et~al. 2025.
\newblock Deepseek-r1: Incentivizing reasoning capability in llms via reinforcement learning.
\newblock \emph{arXiv preprint arXiv:2501.12948}.

\bibitem[{Huang et~al.(2023)Huang, Yong, Ma, Linghu, Li, Wang, Li, Zhu, Jia, and Huang}]{huang2023embodied}
Jiangyong Huang, Silong Yong, Xiaojian Ma, Xiongkun Linghu, Puhao Li, Yan Wang, Qing Li, Song-Chun Zhu, Baoxiong Jia, and Siyuan Huang. 2023.
\newblock An embodied generalist agent in 3d world.
\newblock \emph{arXiv preprint arXiv:2311.12871}.

\bibitem[{Huang et~al.(2024)Huang, Zou, Li, Liu, Zheng, Chern, Xia, Qin, Yuan, and Liu}]{huang2024o1}
Zhen Huang, Haoyang Zou, Xuefeng Li, Yixiu Liu, Yuxiang Zheng, Ethan Chern, Shijie Xia, Yiwei Qin, Weizhe Yuan, and Pengfei Liu. 2024.
\newblock O1 replication journey--part 2: Surpassing o1-preview through simple distillation, big progress or bitter lesson?
\newblock \emph{arXiv preprint arXiv:2411.16489}.

\bibitem[{Inoue et~al.(2024)Inoue, Song, and Fu}]{inoue2024drugagent}
Yoshitaka Inoue, Tianci Song, and Tianfan Fu. 2024.
\newblock Drugagent: Explainable drug repurposing agent with large language model-based reasoning.
\newblock \emph{arXiv preprint arXiv:2408.13378}.

\bibitem[{Jin et~al.(2025)Jin, Zeng, Yue, Wang, Zamani, and Han}]{jin2025search}
Bowen Jin, Hansi Zeng, Zhenrui Yue, Dong Wang, Hamed Zamani, and Jiawei Han. 2025.
\newblock Search-r1: Training llms to reason and leverage search engines with reinforcement learning.
\newblock \emph{arXiv preprint arXiv:2503.09516}.

\bibitem[{Kang et~al.(2025)Kang, Jeong, and Cho}]{kang2025t1}
Minki Kang, Jongwon Jeong, and Jaewoong Cho. 2025.
\newblock T1: Tool-integrated self-verification for test-time compute scaling in small language models.
\newblock \emph{arXiv preprint arXiv:2504.04718}.

\bibitem[{Kaufmann et~al.(2023)Kaufmann, Weng, Bengs, and H{\"u}llermeier}]{kaufmann2023survey}
Timo Kaufmann, Paul Weng, Viktor Bengs, and Eyke H{\"u}llermeier. 2023.
\newblock A survey of reinforcement learning from human feedback.
\newblock \emph{arXiv preprint arXiv:2312.14925}.

\bibitem[{Kumar et~al.(2025)Kumar, Ashraf, Thawakar, Anwer, Cholakkal, Shah, Yang, Torr, Khan, and Khan}]{kumar2025llmreasoningsurvey}
Komal Kumar, Tajamul Ashraf, Omkar Thawakar, Rao~Muhammad Anwer, Hisham Cholakkal, Mubarak Shah, Ming-Hsuan Yang, Phillip~HS Torr, Salman Khan, and Fahad~Shahbaz Khan. 2025.
\newblock Llm post-training: A deep dive into reasoning large language models.
\newblock \emph{arXiv preprint arXiv:2502.21321}.

\bibitem[{Li et~al.(2023)Li, Zhao, Yu, Song, Li, Yu, Li, Huang, and Li}]{li2023api}
Minghao Li, Yingxiu Zhao, Bowen Yu, Feifan Song, Hangyu Li, Haiyang Yu, Zhoujun Li, Fei Huang, and Yongbin Li. 2023.
\newblock Api-bank: A comprehensive benchmark for tool-augmented llms.
\newblock In \emph{Proceedings of the 2023 Conference on Empirical Methods in Natural Language Processing}, pages 3102--3116.

\bibitem[{Li et~al.(2025{\natexlab{a}})Li, Zou, and Liu}]{li2025limr}
Xuefeng Li, Haoyang Zou, and Pengfei Liu. 2025{\natexlab{a}}.
\newblock Limr: Less is more for rl scaling.
\newblock \emph{arXiv preprint arXiv:2502.11886}.

\bibitem[{Li et~al.(2025{\natexlab{b}})Li, Zou, and Liu}]{li2025torl}
Xuefeng Li, Haoyang Zou, and Pengfei Liu. 2025{\natexlab{b}}.
\newblock Torl: Scaling tool-integrated rl.
\newblock \emph{arXiv preprint arXiv:2503.23383}.

\bibitem[{Liao et~al.(2024)Liao, Luo, Li, Wu, and Fan}]{liao2024mario}
Minpeng Liao, Wei Luo, Chengxi Li, Jing Wu, and Kai Fan. 2024.
\newblock Mario: Math reasoning with code interpreter output--a reproducible pipeline.
\newblock \emph{arXiv preprint arXiv:2401.08190}.

\bibitem[{Lin et~al.(2024)Lin, Wen, Peng, Nie, Liao, Wang, Mo, Zhou, Cheng, Zhao et~al.}]{lin2024hammer}
Qiqiang Lin, Muning Wen, Qiuying Peng, Guanyu Nie, Junwei Liao, Jun Wang, Xiaoyun Mo, Jiamu Zhou, Cheng Cheng, Yin Zhao, et~al. 2024.
\newblock Hammer: Robust function-calling for on-device language models via function masking.
\newblock \emph{arXiv preprint arXiv:2410.04587}.

\bibitem[{Ling et~al.(2023)Ling, Zhao, Lu, Deng, Zheng, Wang, Chowdhury, Li, Cui, Zhang et~al.}]{ling2023domain}
Chen Ling, Xujiang Zhao, Jiaying Lu, Chengyuan Deng, Can Zheng, Junxiang Wang, Tanmoy Chowdhury, Yun Li, Hejie Cui, Xuchao Zhang, et~al. 2023.
\newblock Domain specialization as the key to make large language models disruptive: A comprehensive survey.
\newblock \emph{arXiv preprint arXiv:2305.18703}.

\bibitem[{Liu et~al.(2024)Liu, Huang, Zeng, Hao, Yu, Li, Wang, Gan, Liu, Yu et~al.}]{liu2024toolace}
Weiwen Liu, Xu~Huang, Xingshan Zeng, Xinlong Hao, Shuai Yu, Dexun Li, Shuai Wang, Weinan Gan, Zhengying Liu, Yuanqing Yu, et~al. 2024.
\newblock Toolace: Winning the points of llm function calling.
\newblock \emph{arXiv preprint arXiv:2409.00920}.

\bibitem[{Meng et~al.(2024)Meng, Xia, and Chen}]{meng2024simpo}
Yu~Meng, Mengzhou Xia, and Danqi Chen. 2024.
\newblock Simpo: Simple preference optimization with a reference-free reward.
\newblock \emph{Advances in Neural Information Processing Systems}, 37:124198--124235.

\bibitem[{Patil et~al.(2023)Patil, Zhang, Wang, and Gonzalez}]{patil2023gorilla}
Shishir~G Patil, Tianjun Zhang, Xin Wang, and Joseph~E Gonzalez. 2023.
\newblock Gorilla: Large language model connected with massive apis.
\newblock \emph{arXiv preprint arXiv:2305.15334}.

\bibitem[{Patil et~al.(2024)Patil, Zhang, Wang, and Gonzalez}]{patil2024gorilla}
Shishir~G Patil, Tianjun Zhang, Xin Wang, and Joseph~E Gonzalez. 2024.
\newblock Gorilla: Large language model connected with massive apis.
\newblock \emph{Advances in Neural Information Processing Systems}, 37:126544--126565.

\bibitem[{Press et~al.(2022)Press, Zhang, Min, Schmidt, Smith, and Lewis}]{press2022measuring}
Ofir Press, Muru Zhang, Sewon Min, Ludwig Schmidt, Noah~A Smith, and Mike Lewis. 2022.
\newblock Measuring and narrowing the compositionality gap in language models.
\newblock \emph{arXiv preprint arXiv:2210.03350}.

\bibitem[{Qian et~al.(2025)Qian, Acikgoz, Wang, Chen, Sil, Hakkani-T{\"u}r, Tur, and Ji}]{qian2025smart}
Cheng Qian, Emre~Can Acikgoz, Hongru Wang, Xiusi Chen, Avirup Sil, Dilek Hakkani-T{\"u}r, Gokhan Tur, and Heng Ji. 2025.
\newblock Smart: Self-aware agent for tool overuse mitigation.
\newblock \emph{arXiv preprint arXiv:2502.11435}.

\bibitem[{Qian et~al.(2023)Qian, Han, Fung, Qin, Liu, and Ji}]{qian2023creator}
Cheng Qian, Chi Han, Yi~Fung, Yujia Qin, Zhiyuan Liu, and Heng Ji. 2023.
\newblock Creator: Tool creation for disentangling abstract and concrete reasoning of large language models.
\newblock In \emph{Findings of the Association for Computational Linguistics: EMNLP 2023}, pages 6922--6939.

\bibitem[{Qian et~al.(2024{\natexlab{a}})Qian, Han, Luo, He, Chen, Zhang, Du, Yao, Yang, Zhang et~al.}]{qian2024escapebench}
Cheng Qian, Peixuan Han, Qinyu Luo, Bingxiang He, Xiusi Chen, Yuji Zhang, Hongyi Du, Jiarui Yao, Xiaocheng Yang, Denghui Zhang, et~al. 2024{\natexlab{a}}.
\newblock Escapebench: Pushing language models to think outside the box.
\newblock \emph{arXiv preprint arXiv:2412.13549}.

\bibitem[{Qian et~al.(2024{\natexlab{b}})Qian, Xiong, Liu, and Liu}]{qian2024toolink}
Cheng Qian, Chenyan Xiong, Zhenghao Liu, and Zhiyuan Liu. 2024{\natexlab{b}}.
\newblock Toolink: Linking toolkit creation and using through chain-of-solving on open-source model.
\newblock In \emph{Proceedings of the 2024 Conference of the North American Chapter of the Association for Computational Linguistics: Human Language Technologies (Volume 1: Long Papers)}, pages 831--854.

\bibitem[{Qin et~al.(2024{\natexlab{a}})Qin, Li, Zou, Liu, Xia, Huang, Ye, Yuan, Liu, Li et~al.}]{qin2024o1}
Yiwei Qin, Xuefeng Li, Haoyang Zou, Yixiu Liu, Shijie Xia, Zhen Huang, Yixin Ye, Weizhe Yuan, Hector Liu, Yuanzhi Li, et~al. 2024{\natexlab{a}}.
\newblock O1 replication journey: A strategic progress report--part 1.
\newblock \emph{arXiv preprint arXiv:2410.18982}.

\bibitem[{Qin et~al.(2023)Qin, Hu, Lin, Chen, Ding, Cui, Zeng, Huang, Xiao, Han et~al.}]{qin2023tool}
Yujia Qin, Shengding Hu, Yankai Lin, Weize Chen, Ning Ding, Ganqu Cui, Zheni Zeng, Yufei Huang, Chaojun Xiao, Chi Han, et~al. 2023.
\newblock Tool learning with foundation models.
\newblock \emph{arXiv preprint arXiv.2304.08354}, 10.

\bibitem[{Qin et~al.(2024{\natexlab{b}})Qin, Hu, Lin, Chen, Ding, Cui, Zeng, Zhou, Huang, Xiao et~al.}]{qin2024tool}
Yujia Qin, Shengding Hu, Yankai Lin, Weize Chen, Ning Ding, Ganqu Cui, Zheni Zeng, Xuanhe Zhou, Yufei Huang, Chaojun Xiao, et~al. 2024{\natexlab{b}}.
\newblock Tool learning with foundation models.
\newblock \emph{ACM Computing Surveys}, 57(4):1--40.

\bibitem[{Qin et~al.(2024{\natexlab{c}})Qin, Liang, Ye, Zhu, Yan, Lu, Lin, Cong, Tang, Qian, Zhao, Hong, Tian, Xie, Zhou, Gerstein, Li, Liu, and Sun}]{qin2023toolllm}
Yujia Qin, Shihao Liang, Yining Ye, Kunlun Zhu, Lan Yan, Yaxi Lu, Yankai Lin, Xin Cong, Xiangru Tang, Bill Qian, Sihan Zhao, Lauren Hong, Runchu Tian, Ruobing Xie, Jie Zhou, Mark Gerstein, Dahai Li, Zhiyuan Liu, and Maosong Sun. 2024{\natexlab{c}}.
\newblock Toolllm: Facilitating large language models to master 16000+ real-world apis.
\newblock In \emph{The Twelfth International Conference on Learning Representations}.

\bibitem[{Rafailov et~al.(2023)Rafailov, Sharma, Mitchell, Manning, Ermon, and Finn}]{rafailov2023direct}
Rafael Rafailov, Archit Sharma, Eric Mitchell, Christopher~D Manning, Stefano Ermon, and Chelsea Finn. 2023.
\newblock Direct preference optimization: Your language model is secretly a reward model.
\newblock \emph{Advances in Neural Information Processing Systems}, 36:53728--53741.

\bibitem[{Roohani et~al.(2024)Roohani, Lee, Huang, Vora, Steinhart, Huang, Marson, Liang, and Leskovec}]{roohani2024biodiscoveryagent}
Yusuf Roohani, Andrew Lee, Qian Huang, Jian Vora, Zachary Steinhart, Kexin Huang, Alexander Marson, Percy Liang, and Jure Leskovec. 2024.
\newblock Biodiscoveryagent: An ai agent for designing genetic perturbation experiments.
\newblock \emph{arXiv preprint arXiv:2405.17631}.

\bibitem[{Schick et~al.(2023)Schick, Dwivedi-Yu, Dess{\`\i}, Raileanu, Lomeli, Hambro, Zettlemoyer, Cancedda, and Scialom}]{schick2023toolformer}
Timo Schick, Jane Dwivedi-Yu, Roberto Dess{\`\i}, Roberta Raileanu, Maria Lomeli, Eric Hambro, Luke Zettlemoyer, Nicola Cancedda, and Thomas Scialom. 2023.
\newblock Toolformer: Language models can teach themselves to use tools.
\newblock \emph{Advances in Neural Information Processing Systems}, 36:68539--68551.

\bibitem[{Schulman et~al.(2017)Schulman, Wolski, Dhariwal, Radford, and Klimov}]{schulman2017proximal}
John Schulman, Filip Wolski, Prafulla Dhariwal, Alec Radford, and Oleg Klimov. 2017.
\newblock Proximal policy optimization algorithms.
\newblock \emph{arXiv preprint arXiv:1707.06347}.

\bibitem[{Shao et~al.(2024)Shao, Wang, Zhu, Xu, Song, Bi, Zhang, Zhang, Li, Wu et~al.}]{shao2024deepseekmath}
Zhihong Shao, Peiyi Wang, Qihao Zhu, Runxin Xu, Junxiao Song, Xiao Bi, Haowei Zhang, Mingchuan Zhang, YK~Li, Y~Wu, et~al. 2024.
\newblock Deepseekmath: Pushing the limits of mathematical reasoning in open language models.
\newblock \emph{arXiv preprint arXiv:2402.03300}.

\bibitem[{Shen et~al.(2025)Shen, Liu, Li, Fang, Ma, Liao, Shen, Zhang, Zhao, Zhang et~al.}]{shen2025vlm}
Haozhan Shen, Peng Liu, Jingcheng Li, Chunxin Fang, Yibo Ma, Jiajia Liao, Qiaoli Shen, Zilun Zhang, Kangjia Zhao, Qianqian Zhang, et~al. 2025.
\newblock Vlm-r1: A stable and generalizable r1-style large vision-language model.
\newblock \emph{arXiv preprint arXiv:2504.07615}.

\bibitem[{Sheng et~al.(2024)Sheng, Zhang, Ye, Wu, Zhang, Zhang, Peng, Lin, and Wu}]{sheng2024hybridflow}
Guangming Sheng, Chi Zhang, Zilingfeng Ye, Xibin Wu, Wang Zhang, Ru~Zhang, Yanghua Peng, Haibin Lin, and Chuan Wu. 2024.
\newblock Hybridflow: A flexible and efficient rlhf framework.
\newblock \emph{arXiv preprint arXiv:2409.19256}.

\bibitem[{Song et~al.(2025)Song, Jiang, Min, Chen, Chen, Zhao, Fang, and Wen}]{song2025r1}
Huatong Song, Jinhao Jiang, Yingqian Min, Jie Chen, Zhipeng Chen, Wayne~Xin Zhao, Lei Fang, and Ji-Rong Wen. 2025.
\newblock R1-searcher: Incentivizing the search capability in llms via reinforcement learning.
\newblock \emph{arXiv preprint arXiv:2503.05592}.

\bibitem[{Team et~al.(2025)Team, Du, Gao, Xing, Jiang, Chen, Li, Xiao, Du, Liao et~al.}]{team2025kimi}
Kimi Team, Angang Du, Bofei Gao, Bowei Xing, Changjiu Jiang, Cheng Chen, Cheng Li, Chenjun Xiao, Chenzhuang Du, Chonghua Liao, et~al. 2025.
\newblock Kimi k1. 5: Scaling reinforcement learning with llms.
\newblock \emph{arXiv preprint arXiv:2501.12599}.

\bibitem[{Team(2024)}]{2024qwen2.5}
Qwen Team. 2024.
\newblock \href {https://qwenlm.github.io/blog/qwen2.5/} {Qwen2.5: A party of foundation models}.

\bibitem[{Vu et~al.(2023)Vu, Iyyer, Wang, Constant, Wei, Wei, Tar, Sung, Zhou, Le et~al.}]{vu2023freshllms}
Tu~Vu, Mohit Iyyer, Xuezhi Wang, Noah Constant, Jerry Wei, Jason Wei, Chris Tar, Yun-Hsuan Sung, Denny Zhou, Quoc Le, et~al. 2023.
\newblock Freshllms: Refreshing large language models with search engine augmentation.
\newblock \emph{arXiv preprint arXiv:2310.03214}.

\bibitem[{Wang et~al.(2024)Wang, Guo, Yao, Zhang, Zhang, Wu, Zhang, Dai, Wen, Ye et~al.}]{wang2024autosurvey}
Yidong Wang, Qi~Guo, Wenjin Yao, Hongbo Zhang, Xin Zhang, Zhen Wu, Meishan Zhang, Xinyu Dai, Qingsong Wen, Wei Ye, et~al. 2024.
\newblock Autosurvey: Large language models can automatically write surveys.
\newblock \emph{Advances in Neural Information Processing Systems}, 37:115119--115145.

\bibitem[{Xie et~al.(2025)Xie, Gao, Ren, Luo, Hong, Dai, Zhou, Qiu, Wu, and Luo}]{xie2025logic}
Tian Xie, Zitian Gao, Qingnan Ren, Haoming Luo, Yuqian Hong, Bryan Dai, Joey Zhou, Kai Qiu, Zhirong Wu, and Chong Luo. 2025.
\newblock Logic-rl: Unleashing llm reasoning with rule-based reinforcement learning.
\newblock \emph{arXiv preprint arXiv:2502.14768}.

\bibitem[{Yao et~al.(2023)Yao, Zhao, Yu, Du, Shafran, Narasimhan, and Cao}]{yao2023react}
Shunyu Yao, Jeffrey Zhao, Dian Yu, Nan Du, Izhak Shafran, Karthik~R Narasimhan, and Yuan Cao. 2023.
\newblock React: Synergizing reasoning and acting in language models.
\newblock In \emph{The Eleventh International Conference on Learning Representations}.

\bibitem[{Ye et~al.(2023)Ye, Cong, Tian, Qin, Liu, Lin, Liu, and Sun}]{ye2023rational}
Yining Ye, Xin Cong, Shizuo Tian, Yujia Qin, Chong Liu, Yankai Lin, Zhiyuan Liu, and Maosong Sun. 2023.
\newblock Rational decision-making agent with internalized utility judgment.
\newblock \emph{arXiv preprint arXiv:2308.12519}.

\bibitem[{Yu et~al.(2025)Yu, Zhang, Zhu, Yuan, Zuo, Yue, Fan, Liu, Liu, Liu et~al.}]{yu2025dapo}
Qiying Yu, Zheng Zhang, Ruofei Zhu, Yufeng Yuan, Xiaochen Zuo, Yu~Yue, Tiantian Fan, Gaohong Liu, Lingjun Liu, Xin Liu, et~al. 2025.
\newblock Dapo: An open-source llm reinforcement learning system at scale.
\newblock \emph{arXiv preprint arXiv:2503.14476}.

\bibitem[{Yu et~al.(2024)Yu, Wang, Ma, Guo, Zhan, Wang, Wu, Guo, and Zhang}]{yu2024steptool}
Yuanqing Yu, Zhefan Wang, Weizhi Ma, Zhicheng Guo, Jingtao Zhan, Shuai Wang, Chuhan Wu, Zhiqiang Guo, and Min Zhang. 2024.
\newblock Steptool: A step-grained reinforcement learning framework for tool learning in llms.
\newblock \emph{arXiv preprint arXiv:2410.07745}.

\bibitem[{Yuan et~al.(2025)Yuan, Yu, Zuo, Zhu, Xu, Chen, Wang, Fan, Du, Wei et~al.}]{yuan2025vapo}
Yufeng Yuan, Qiying Yu, Xiaochen Zuo, Ruofei Zhu, Wenyuan Xu, Jiaze Chen, Chengyi Wang, TianTian Fan, Zhengyin Du, Xiangpeng Wei, et~al. 2025.
\newblock Vapo: Efficient and reliable reinforcement learning for advanced reasoning tasks.
\newblock \emph{arXiv preprint arXiv:2504.05118}.

\bibitem[{Zeng et~al.(2024)Zeng, Liu, Lu, Wang, Liu, Dong, and Tang}]{zeng2024agenttuning}
Aohan Zeng, Mingdao Liu, Rui Lu, Bowen Wang, Xiao Liu, Yuxiao Dong, and Jie Tang. 2024.
\newblock \href {https://doi.org/10.18653/v1/2024.findings-acl.181} {{A}gent{T}uning: Enabling generalized agent abilities for {LLM}s}.
\newblock In \emph{Findings of the Association for Computational Linguistics: ACL 2024}, pages 3053--3077, Bangkok, Thailand. Association for Computational Linguistics.

\bibitem[{Zhai et~al.(2024)Zhai, Yang, Xu, Dawei, Yang, Ding, and Wang}]{zhai2024enhancing}
Yuanzhao Zhai, Tingkai Yang, Kele Xu, Feng Dawei, Cheng Yang, Bo~Ding, and Huaimin Wang. 2024.
\newblock Enhancing decision-making for llm agents via step-level q-value models.
\newblock \emph{arXiv preprint arXiv:2409.09345}.

\bibitem[{Zhang et~al.(2023)Zhang, Du, Shan, Zhou, Du, Tenenbaum, Shu, and Gan}]{zhang2023building}
Hongxin Zhang, Weihua Du, Jiaming Shan, Qinhong Zhou, Yilun Du, Joshua~B Tenenbaum, Tianmin Shu, and Chuang Gan. 2023.
\newblock Building cooperative embodied agents modularly with large language models.
\newblock \emph{arXiv preprint arXiv:2307.02485}.

\bibitem[{Zhang et~al.(2024)Zhang, Lan, Zhu, Liu, Hoang, Kokane, Yao, Tan, Prabhakar, Chen et~al.}]{zhang2024xlam}
Jianguo Zhang, Tian Lan, Ming Zhu, Zuxin Liu, Thai Hoang, Shirley Kokane, Weiran Yao, Juntao Tan, Akshara Prabhakar, Haolin Chen, et~al. 2024.
\newblock xlam: A family of large action models to empower ai agent systems.
\newblock \emph{arXiv preprint arXiv:2409.03215}.

\bibitem[{Zheng et~al.(2025)Zheng, Fu, Hu, Cai, Ye, Lu, and Liu}]{zheng2025deepresearcher}
Yuxiang Zheng, Dayuan Fu, Xiangkun Hu, Xiaojie Cai, Lyumanshan Ye, Pengrui Lu, and Pengfei Liu. 2025.
\newblock Deepresearcher: Scaling deep research via reinforcement learning in real-world environments.
\newblock \emph{arXiv preprint arXiv:2504.03160}.

\end{thebibliography}

\clearpage
\appendix

\section*{Appendix}
\label{sec:appendix}

\begin{figure*}[t]
\centering
\resizebox{0.85\textwidth}{!}{
\begin{tcolorbox}[colback=gray!5!white, colframe=blue!75!black, 
title=User Prompt for Training, boxrule=0.3mm, width=\textwidth, arc=3mm, auto outer arc=true]
\textbf{Dialogue History}\\
\textless{}user\textgreater{} \{\{ Initial User Input \}\} \textless{}/user\textgreater{}\\
\\
\textcolor{deepblue}{\textless{}think\textgreater{}} Round 1 Model Thought \textcolor{deepblue}{\textless{}/think\textgreater{}}\\
\{\{ Round 1 model output \textcolor{deepgreen}{\textless{}tool\_call\textgreater{}} or \textcolor{deeppurple}{\textless{}response\textgreater{}} \}\}\\
\textcolor{brown}{\textless{}obs\textgreater{}} Round 1 Observation \textcolor{brown}{\textless{}/obs\textgreater{}}\\
... ...\\
\\
\textless{}user\textgreater{} \{\{ User Input \}\} \textless{}/user\textgreater{}\\
... ...\\
\end{tcolorbox}
}
\caption{The user prompt used for TIR's rollout.}
\label{prompt:user}
\end{figure*}

\section{User Prompt Details}
The system instruction is shown in \Cref{prompt:system}. The user prompt is used to store the trajectory history, including intermediate thoughts, tool calls, environment observations, and any additional user commands. The complete user instruction is presented in \Cref{prompt:user}.

\section{Experiment Details}
\label{apdx:training}

\paragraph{Training Data Details.} We empirically use 4K data points for training, as each dataset consists of samples drawn from the same distribution. Adding more data of similar nature does not increase task diversity. Moreover, we observe that increasing the dataset size beyond 4K does not yield noticeable improvements in the training convergence or final performance, suggesting diminishing returns from additional data under this setting.

\paragraph{GRPO Setting Details.} For all the tool calls in the dataset, we all use JSON format to represent tool call as it's easy to parse and is the most general and structure way of performing tool call. For the GRPO training, we use 2 A100 (80G) GPUs per run with the following hyper-parameters:

\begin{table}[ht]
\centering
\small
\begin{tabular}{ll}
\toprule
\textbf{Category} & \textbf{Hyperparameter} \\
\midrule
\multicolumn{2}{l}{\textbf{Data Configuration}} \\
\midrule
Train Batch Size & 512 \\
Validation Batch Size & 128 \\
Max Prompt Length & 2048 \\
Max Response Length & 1024 \\
\midrule
\multicolumn{2}{l}{\textbf{Optimization}} \\
\midrule
Learning Rate & 1e-6 \\
PPO Mini Batch Size & 128 \\
KL Loss Used & False \\
\midrule
\multicolumn{2}{l}{\textbf{Rollout Configuration}} \\
\midrule
Rollout Name & vllm \\
GPU Memory Utilization & 0.6 \\
Number of Rollouts & 4 \\
\midrule
\multicolumn{2}{l}{\textbf{Training \& Logging}} \\
\midrule
Save Frequency (Steps) & 15 \\
Test Frequency (Steps) & 5 \\
Total Epochs & 15 \\
\bottomrule
\end{tabular}
\caption{Configuration for GRPO training.}
\label{tab:training_hparams}
\end{table}

\paragraph{Baselines.} The 400 selected data points used for SFT share the same distribution as the 4k data points used for RL training, but differ in content. For SFT, each data point includes a \textcolor{deepblue}{\textless{}think\textgreater{}} field, with thought content distilled from Deepseek-R1 trajectories. In contrast, GRPO does not require ground truth thought, as only the tool calls are used to compute rewards in the GRPO setting.

We use 400 data points for SFT based on empirical observations that this amount is sufficient to help the raw model learn to follow our tool call format. This provides a stronger initialization and reduces the burden of learning the format from scratch during RL training. However, we also find that relying solely on SFT can lead to overfitting, which may ultimately degrade performance.

\paragraph{PPO Setting Details.} We apply approximately the same parameter settings as GRPO for the PPO training. Similarly, we use 2 A100 (80G) GPUs per run with the following hyper-parameters:

\begin{table}[ht]
\centering
\small
\begin{tabular}{ll}
\toprule
\textbf{Category} & \textbf{Hyperparameter} \\
\midrule
\multicolumn{2}{l}{\textbf{Data Configuration}} \\
\midrule
Train Batch Size & 512 \\
Validation Batch Size & 128 \\
Max Prompt Length & 1024 \\
Max Response Length & 512 \\
\midrule
\multicolumn{2}{l}{\textbf{Optimization}} \\
\midrule
Actor Learning Rate & 1e-6 \\
Critic Learning Rate & 1e-5 \\
PPO Mini Batch Size & 128 \\
PPO Micro Batch Size & 8 \\
KL Coefficient & 0.001 \\
\midrule
\multicolumn{2}{l}{\textbf{Rollout Configuration}} \\
\midrule
Rollout Name & vllm \\
GPU Memory Utilization & 0.3 \\
\midrule
\multicolumn{2}{l}{\textbf{Training \& Logging}} \\
\midrule
Critic Warmup Steps & 0 \\
Save Frequency (Steps) & 15 \\
Test Frequency (Steps) & 5 \\
Total Epochs & 15 \\
\bottomrule
\end{tabular}
\caption{Configuration for PPO training.}
\label{tab:training_hparams2}
\end{table}

\section{Additional Results}

We present additional results on three benchmarks, applying GRPO and PPO methods to models initialized with SFT on 4K data points. This setting serves as a ``theoretical'' upper bound, since the same 4K data is first used for SFT and subsequently reused for RL training.

\begin{table*}[!t]
\centering
\resizebox{1.0\linewidth}{!}{
\begin{tabular}{lccccccc}
\toprule
\textbf{Model} & \textbf{Overall Acc} & Non-Live AST Acc & Non-Live Exec Acc & Live Acc & Multi Turn Acc & Relevance Detection & Irrelevance Detection \\
\midrule
Qwen2.5-1.5B-Instruct (\textbf{Raw}) & 19.41\% & 16.00\% & 13.18\% & 35.58\% & 0.00\% & 44.44\% & 82.49\% \\
Qwen2.5-1.5B-Instruct (\textbf{SFT400+PPO}) & 42.95\% & 77.65\% & 69.75\% & 55.73\% & 1.88\% & 100.00\% & 48.40\% \\
Qwen2.5-1.5B-Instruct (\textbf{SFT400+GRPO}) & 40.93\% & 70.54\% & 60.79\% & 56.33\% & 1.00\% & 94.44\% & 58.63\% \\
Qwen2.5-1.5B-Instruct (\textbf{SFT4k+PPO}) & 40.24\% & 66.42\% & 62.02\% & 54.58\% & 2.50\% & 94.12\% & 55.09\% \\
Qwen2.5-1.5B-Instruct (\textbf{SFT4k+GRPO}) & 42.63\% & 66.60\% & 64.77\% & 60.15\% & 1.38\% & 88.89\% & 67.98\% \\
\midrule
Qwen2.5-3B-Instruct (\textbf{Raw}) & 33.04\% & 42.52\% & 40.80\% & 53.96\% & 1.00\% & 64.71\% & 56.01\% \\
Qwen2.5-3B-Instruct (\textbf{SFT400+PPO}) & 45.80\% & 78.29\% & 71.09\% & 58.76\% & 5.12\% & 94.12\% & 54.70\% \\
Qwen2.5-3B-Instruct (\textbf{SFT400+GRPO}) & 46.42\% & 76.21\% & 68.93\% & 64.15\% & 1.75\% & 88.89\% & 71.76\% \\
Qwen2.5-3B-Instruct (\textbf{SFT4k+PPO}) & 48.22\% & 77.75\% & 73.18\% & 64.27\% & 5.25\% & 94.12\% & 66.41\% \\
Qwen2.5-3B-Instruct (\textbf{SFT4k+GRPO}) & 47.82\% & 75.12\% & 69.52\% & 68.19\% & 2.38\% & 77.78\% & 76.16\% \\
\midrule
Qwen2.5-7B-Instruct (\textbf{Raw}) & 41.97\% & 66.02\% & 70.11\% & 53.51\% & 4.25\% & 76.47\% & 62.66\% \\
Qwen2.5-7B-Instruct (\textbf{SFT400+PPO}) & 42.02\% & 83.90\% & 72.62\% & 51.84\% & 0.25\% & 100\% & 29.66\% \\
Qwen2.5-7B-Instruct (\textbf{SFT400+GRPO}) & 39.25\% & 80.69\% & 74.34\% & 46.51\% & 0.25\% & 100\% & 14.19\% \\
Qwen2.5-7B-Instruct (\textbf{SFT4k+PPO}) & 33.80\% & 42.67\% & 49.50\% & 51.80\% & 2.38\% & 77.78\% & 55.79\% \\
Qwen2.5-7B-Instruct (\textbf{SFT4k+GRPO}) & 35.18\% & 43.58\% & 50.39\% & 55.49\% & 0.87\% & 77.78\% & 67.12\% \\
\midrule
Llama-3.2-3B-Instruct (\textbf{Raw}) & 22.09\% & 17.44\% & 14.57\% & 43.85\% & 0.00\% & 77.78\% & 66.07\% \\
Llama-3.2-3B-Instruct (\textbf{SFT400+PPO}) & 41.62\% & 68.10\% & 69.88\% & 52.98\% & 3.00\% & 94.12\% & 56.29\% \\
Llama-3.2-3B-Instruct (\textbf{SFT400+GRPO}) & 42.54\% & 65.15\% & 68.98\% & 59.40\% & 0.88\% & 72.22\% & 65.80\% \\
Llama-3.2-3B-Instruct (\textbf{SFT4k+PPO}) & 45.41\% & 73.71\% & 68.46\% & 62.27\% & 2.50\% & 82.35\% & 68.75\% \\
Llama-3.2-3B-Instruct (\textbf{SFT4k+GRPO}) & 45.50\% & 70.69\% & 67.70\% & 64.73\% & 1.00\% & 77.78\% & 78.85\% \\
\bottomrule
\end{tabular}
}
\caption{BFCL V3 Benchmark Results (Additional Result)}
\label{tab:bfcl-result-additional}
\end{table*}


\begin{table*}[!t]
\centering

\begin{minipage}[t]{0.51\linewidth}
\centering

\resizebox{1\linewidth}{!}{
\begin{tabular}{lcccc}
\toprule
\textbf{Model} & \textbf{Overall Acc} & Level 1 & Level 2 & Level 3 \\
\midrule
Qwen2.5-1.5B-Instruct (\textbf{Raw}) & 30.65\% & 28.32\% & 35.82\% & 35.11\% \\
Qwen2.5-1.5B-Instruct (\textbf{SFT400+PPO}) & 57.12\% & 60.9\% & 50.75\% & 48.85\% \\
Qwen2.5-1.5B-Instruct (\textbf{SFT400+GRPO}) & 61.31\% & 64.16\% & 58.21\% & 54.20\% \\
Qwen2.5-1.5B-Instruct (\textbf{SFT4k+PPO}) & 61.31\% & 64.91\% & 56.72\% & 52.67\% \\
Qwen2.5-1.5B-Instruct (\textbf{SFT4k+GRPO}) & 59.46\% & 65.16\% & 53.73\% & 45.04\% \\
\midrule
Qwen2.5-3B-Instruct (\textbf{Raw}) & 51.59\% & 59.65\% & 32.84\% & 36.64\% \\
Qwen2.5-3B-Instruct (\textbf{SFT400+PPO}) & 65.16\% & 67.92\% & 55.22\% & 61.83\% \\
Qwen2.5-3B-Instruct (\textbf{SFT400+GRPO}) & 62.48\% & 68.67\% & 58.21\% & 45.80\% \\
Qwen2.5-3B-Instruct (\textbf{SFT4k+PPO}) & 60.13\% & 64.41\% & 44.78\% & 54.96\% \\
Qwen2.5-3B-Instruct (\textbf{SFT4k+GRPO}) & 60.80\% & 64.41\% & 56.72\% & 51.91\% \\
\midrule
Qwen2.5-7B-Instruct (\textbf{Raw}) & 62.48\% & 70.68\% & 49.25\% & 44.27\% \\
Qwen2.5-7B-Instruct (\textbf{SFT400+PPO}) & 63.15\% & 72.43\% & 58.21\% & 37.4\% \\
Qwen2.5-7B-Instruct (\textbf{SFT400+GRPO}) & 54.10\% & 61.40\% & 52.24\% &32.82\% \\
Qwen2.5-7B-Instruct (\textbf{SFT4k+PPO}) & 59.30\% & 61.40\% & 40.30\% & 61.60\% \\
Qwen2.5-7B-Instruct (\textbf{SFT4k+GRPO}) & 52.60\% & 56.39\% & 34.33\% & 50.38\% \\
\midrule
Llama-3.2-3B-Instruct (\textbf{Raw}) & 40.54\% & 44.86\% & 29.85\% & 32.82\% \\
Llama-3.2-3B-Instruct (\textbf{SFT400+PPO}) & 57.79\% & 63.16\% & 47.76\% & 46.56\% \\
Llama-3.2-3B-Instruct (\textbf{SFT400+GRPO}) & 56.78\% & 63.60\% & 41.79\% & 43.51\% \\
Llama-3.2-3B-Instruct (\textbf{SFT4k+PPO}) & 54.10\% & 60.65\% & 40.30\% & 41.22\% \\
Llama-3.2-3B-Instruct (\textbf{SFT4k+GRPO}) & 50.92\% & 59.15\% & 34.33\% & 34.35\% \\
\bottomrule
\end{tabular}
}
\caption{API-Bank Test Results (Additional Result)}
\label{tab:apibank-result-additional}

\end{minipage}
\hfill
\begin{minipage}[t]{0.452\textwidth}
\centering

\resizebox{1\linewidth}{!}{
\begin{tabular}{lcccc}
\toprule
\textbf{Model} & \textbf{Accuracy} & Avg Num Tool Call \\
\midrule
Qwen2.5-1.5B-Instruct (\textbf{Raw}) & 20.8\% & 0.61 \\
Qwen2.5-1.5B-Instruct (\textbf{SFT400+PPO}) & 36.8\% & 1.06 \\
Qwen2.5-1.5B-Instruct (\textbf{SFT400+GRPO}) & 38.4\% & 0.96 \\
Qwen2.5-1.5B-Instruct (\textbf{SFT4k+PPO}) & 36.8\% & 1.06 \\
Qwen2.5-1.5B-Instruct (\textbf{SFT4k+GRPO}) & 34.4\% & 1.02 \\
\midrule
Qwen2.5-3B-Instruct (\textbf{Raw}) & 52.0\% & 1.77 \\
Qwen2.5-3B-Instruct (\textbf{SFT400+PPO}) & 43.2\% & 1.04 \\
Qwen2.5-3B-Instruct (\textbf{SFT400+GRPO}) & 56.8\% & 0.99 \\
Qwen2.5-3B-Instruct (\textbf{SFT4k+PPO}) & 46.4\% & 1.01 \\
Qwen2.5-3B-Instruct (\textbf{SFT4k+GRPO}) & 47.2\% & 0.98 \\
\midrule
Qwen2.5-7B-Instruct (\textbf{Raw}) & 69.6\% & 1.42 \\
Qwen2.5-7B-Instruct (\textbf{SFT400+PPO}) & 45.6\% & 3.54 \\
Qwen2.5-7B-Instruct (\textbf{SFT400+GRPO}) & 29.6\% & 3.70 \\
Qwen2.5-7B-Instruct (\textbf{SFT4k+PPO}) & 40.0\% & 1.25 \\
Qwen2.5-7B-Instruct (\textbf{SFT4k+GRPO}) & 32.0\% & 1.25 \\
\midrule
Llama-3.2-3B-Instruct (\textbf{Raw}) & 34.4\% & 1.25 \\
Llama-3.2-3B-Instruct (\textbf{SFT400+PPO}) & 39.2\% & 1.33 \\
Llama-3.2-3B-Instruct (\textbf{SFT400+GRPO}) & 45.6\% & 1.00 \\
Llama-3.2-3B-Instruct (\textbf{SFT4k+PPO}) & 49.6\% & 1.02 \\
Llama-3.2-3B-Instruct (\textbf{SFT4k+GRPO}) & 42.4\% & 1.03 \\
\bottomrule
\end{tabular}
}
\caption{Bamboogle Test Results (Additional Result)}
\label{tab:bamboogle-result-additional}

\end{minipage}

\end{table*}

The results are shown in \Cref{tab:bfcl-result-additional}, \Cref{tab:apibank-result-additional}, and \Cref{tab:bamboogle-result-additional} for BFCL, API-Bank, and Bamboogle, respectively. We compare RL training initialized with models fine-tuned on either 400 or 4K SFT data points.

Interestingly, our findings suggest that initializing from a model finetuned on 4K data does not consistently outperform initialization from a model finetuned on only 400 data points. In the BFCL benchmark, we even observe cases where performance drops below that of the raw instruct model. This counterintuitive result may stem from overfitting during the SFT phase, which could restrict the model’s ability to explore during RL training and lead to poorer generalization on held-out tasks.

\end{document}